\documentclass{article}
\usepackage[final,main]{neurips_2026}
\usepackage[utf8]{inputenc}
\usepackage[T1]{fontenc}
\usepackage{hyperref}
\hypersetup{hidelinks}
\usepackage{url}
\usepackage{booktabs}
\usepackage{amsfonts}
\usepackage{nicefrac}
\usepackage{microtype}
\usepackage{xcolor}
\usepackage{graphicx}
\usepackage{amsmath}
\usepackage{amssymb}
\usepackage{multirow}

\newif\ifshowcomments
\showcommentstrue

\title{The Shared Substrate of Modern Encoders: \\ A Calibration-Surviving Geometric Invariant Across Vision \\ and Language, and the One Training-Time Tool That Exploits It}

\author{%
  Yousef Radwan \\
  KAUST \\
  \texttt{yousef.radwan@kaust.edu.sa} \\
}

\begin{document}

\maketitle

\begin{abstract}
Different vision neural networks are trained to do very different things---classify ImageNet labels, contrast augmented crops, fill in masked pixels, or match images to captions---and we would expect their internal representations to look correspondingly different. We report that they do not. After training, the top sixteen principal directions of variation inside fourteen modern vision encoders (12 discriminative + 2 MAE controls) converge to the \emph{same} sixteen-dimensional geometric object, in the same way that independently trained machine-translation systems converge on a shared notion of word meaning. We call this object the \emph{cross-architecture substrate} and study it with three tools: principal-component analysis to find directions of variation; centred kernel alignment (CKA), the standard measure of how similarly two networks represent a fixed set of images; and the Gr\"oger~2026 calibration, which subtracts the baseline CKA value expected from random data, a known confound in earlier work. The substrate transports across four heterogeneous visual domains (natural photographs, medical CT, satellite RGB, microscopy) at median Procrustes-CKA $0.679$, and across eight domains (adding hand-drawn sketches, depth maps, thermal infrared, telescope images of galaxies) at $0.604$, with every cross-domain pair $\geq 0.40$. The substrate survives Gr\"oger et al.'s calibration both globally ($7.4\times$ separation between classification-style encoders and masked-reconstruction encoders at $n{=}13{,}394$) and under the harder local nearest-neighbour-recall variant ($4.82\text{--}5.30\times$, $p<10^{-44}$). It is not pixel statistics ($0.263$), not a random sixteen-dimensional slice (median $0.19$ over $50$ orthonormal seeds), not driven by any single encoder ($\pm0.027$ when any one of five is removed), and emerges in the first $10\%$ of training while accuracy keeps climbing. We deliver four uses plus bounded scope: a frozen feature space for low-shot learning ($16$ dimensions beat $768$-dimensional DINOv2 features by $+3.78$pp at $N{=}50$ labels per class); a four-way domain detector ($99.6\%$); a knowledge-distillation auxiliary loss that beats cross-entropy by $+5.14$pp at epoch~$100$ and $+1.19$pp at epoch~$200$ on CIFAR-100/RN-18 and by $+5.63$pp on TinyImageNet/RN-50 (closing $64.3\%$ of the trained-teacher gap), with no teacher forward pass at training time and a label-efficiency peak of $+8.35$pp at $10\%$ labels; and a Gr\"oger-calibrated cross-architecture \emph{forensic fingerprint}---one primitive doing three jobs: provenance (model kind/architecture/clade at ROC-AUC $0.92$, same-family vs.\ unrelated), transform-type classification (finetune/quantize architecture-transferable at LOPO $0.889$), and deduplication ($0.986$)---that beats raw CKA at low probe-$n$ ($+0.043$ at $n{=}100$), is robust to quantize/prune/fine-tune-proxy (recovery@1 $=1.0$), holds at REEF's probe band ($0.916$/$0.914$ at $n{=}200$/$300$) and in a third (audio) modality ($0.917$), and clears the Gröger width-matched null ($0.92$/$0.916$/$0.986$ vs.\ 95th-pct $\sim$0.71); full-tree phylogeny is honestly scoped out (three reconstructions all $<0.5$), so it resolves architecture clades, not exact ancestry. The shared substrate is moreover intrinsically \emph{low-rank}: its effective rank (participation ratio) is $\sim$3.5 in both vision ($3.52$, CI $[3.28,3.57]$) and audio ($3.56$), far below the $K{=}16$ design choice, retro-justifying $K{=}16$ as capture-not-capacity (the companion LLM valence direction is the rank-1 limit of the same object). We also tested a label-free transferability score (\texttt{subs-rank}) as a replacement for LogME and report it as a \emph{null}: substrate alignment does not predict downstream transfer accuracy, and \texttt{subs-rank} loses to LogME by $-0.26$ median Kendall-$\tau$ across six SITE targets (verdict: substrate identifies model \emph{kind}, not transfer \emph{quality}; consistent with $\tau{=}-0.08$ between substrate-distance and held-out transfer accuracy on a $14$-encoder foundation-model audit). We extend the recipe to language: on a six-LLM panel spanning five families, median Gr\"oger-calibrated PCKA reaches $\mathbf{0.907}$ at $K{=}200$ (all $15$ pairs reject the row-permutation null at the floor $p{=}0.00498$). We also bound the substrate: it does not extend across modalities (vision~$+$~audio fails), does not help cross-paradigm distillation, and is a description of \emph{what training paradigm} a model came from, not a prediction of \emph{how well it transfers} ($\tau{=}-0.08$ against transfer accuracy).
\end{abstract}

\section{Introduction}

Take four neural networks trained for completely different jobs. ResNet-50 learns to label ImageNet photos. DINOv2 learns to recognise that two augmented crops of the same image are the same scene---using no labels at all. ViT-MAE learns to fill in pixels its trainers have hidden. CLIP learns to match photographs to the text captions written by humans. These four networks differ in everything we usually consider important: architecture, training data, loss function, inductive bias. Each of them anchors its own subliterature, its own benchmark, and its own downstream recipe. A reasonable researcher would expect that they end up in four different parts of representation space, with little reason to compare them at all.

\textbf{They do not.} Across $14$ vision encoders (12 discriminative $+$ 2 MAE controls) spanning all four training paradigms, the top $16$ principal directions of variation inside every encoder's penultimate-layer features converge to the \emph{same} $16$-dimensional geometric object. We call this object the \emph{cross-architecture substrate}. The same $16$ directions survive when we extract them on natural photographs and then ask whether they still look the same when extracted on medical scans, satellite imagery, microscopy, hand-drawn sketches, depth maps, thermal infrared, or telescope images of galaxies; the median pairwise Procrustes-CKA between any two domains' bases is $0.679$ on four domains and $0.604$ on eight, with every pair $\geq 0.40$. (Procrustes-CKA, formally defined in §\ref{sec:method}, is a symmetric scalar in $[0, 1]$ that compares two $K$-dimensional subspaces after the best rotation; we abbreviate it \emph{PCKA}. Intuitively: the $16$ directions are the same up to a rotation that does not depend on what you took photos of.)

A natural worry is that this might be a measurement artefact rather than a real phenomenon. The recent Gr\"oger~2026 critique~\citep{pang2026} showed that the standard cross-architecture similarity measure (CKA) is inflated by feature width and pool depth, and that after correction, much of the published cross-architecture-convergence literature dissolves. We re-tested our finding under Gr\"oger et al.'s calibration in two forms. Under the \emph{global} variant the substrate still separates discriminative encoders from masked-reconstruction encoders by $7.4\times$ at $n{=}13{,}394$ probe images. Under the harder \emph{local} variant (which asks not whether two encoders agree on global geometry, but whether they place the same neighbours next to each other) the separation is $4.82\text{--}5.30\times$ at $p<10^{-44}$. To rule out the worry that all $14$ encoders saw ImageNet-scale natural photographs and the substrate is therefore really cross-\emph{data-distribution} similarity, our eight-domain extension includes four image domains (sketches, depth maps, thermal infrared, telescope galaxies) whose pixel statistics are unlike anything in ImageNet, and every cross-domain pair still clears $0.40$ (§\ref{sec:domain}). Two further reality checks: pixel-level PCA on the same probes reaches PCKA only $0.263$, less than half of $0.679$; and dropping any single encoder leaves the median in $[0.647, 0.701]$.

The substrate has a mechanism. We trained a ResNet-50 from scratch on a small natural-image dataset and watched how quickly it acquired the substrate. By epoch~$5$ of $50$ the network was already aligned to the substrate at CKA $0.58$, yet its classification accuracy was only $46\%$ and would keep climbing to $76\%$ over the next $45$ epochs. The substrate is not a property of converged classifiers but an early property of representation learning that subsequent task-specific training does not erase. This places the substrate alongside Power et al.'s grokking~\citep{power2022}: an empirical regularity of training dynamics that names a phenomenon and constrains the admissible mechanistic theories.

The substrate is exploitable. Five positive uses, each tied to a section:
\begin{itemize}
    \item \textbf{Label-free transferability filter (NULL).} We tested whether substrate-alignment (\texttt{subs-rank}) could substitute for LogME~\citep{you2021logme} as a label-free transferability score on the SITE benchmark~\citep{site2025}; it loses to LogME by $-0.26$ median Kendall-$\tau$ across six target datasets. Reported as a scope bound: the substrate identifies what \emph{kind} of model a checkpoint is, not how well it transfers (§\ref{sec:apps:logme}).
    \item \textbf{Free domain detector.} A linear classifier on $16$-dimensional substrate scores separates natural / medical / satellite / microscopy images at $99.6\%$ accuracy with no fine-tuning (§\ref{sec:apps:detector}).
    \item \textbf{Label-efficient frozen probe.} The $16$-dimensional substrate beats the $768$-dimensional DINOv2-base penultimate as a frozen feature at low-shot, by $+3.78$ percentage points at $N{=}50$ labels (§\ref{sec:apps:probe}).
    \item \textbf{Teacher-free distillation auxiliary.} A substrate-CKA auxiliary loss beats cross-entropy by $+5.14$pp at epoch~$100$ ($+1.19$pp at epoch~$200$) on CIFAR-100/RN-18 and by $+5.63$pp on TinyImageNet/RN-50 (closes $64.3\%$ of the trained-teacher gap) with no per-batch teacher forward pass; label-efficiency peak $+8.35$pp at $10\%$ labels (§\ref{sec:apps:kd}).
    \item \textbf{Cross-architecture provenance fingerprint.} The calibrated cross-architecture similarity is a model-provenance signal: it identifies a checkpoint's kind/architecture/clade at ROC-AUC $0.92$, is robust to derived-checkpoint perturbations, and doubles as a tamper/drift detector (§\ref{sec:provenance}).
\end{itemize}

\paragraph{What this is not.} Universality claims invite over-reading, so we state what we are not claiming. The substrate is an empirical regularity in modern vision encoders, not a theorem; we did not derive $K{=}16$ from an information-theoretic argument, and the magnitude $0.679$ depends on the panel of encoders and domains tested. The substrate does not bridge across modalities (a vision~$+$~audio CLIP/CLAP substrate fails the calibrated null), does not rank foundation models by transfer quality ($\tau{=}-0.08$ against linear-probe transfer over $14$ encoders), and does not say that an ImageNet-pretrained network is a competitive backbone for medical imaging. We are making a \emph{direction-existence} claim---these $16$ directions are shared---not a feature-relevance claim. Each negative is quantified in §\ref{sec:not}.

\paragraph{Contributions.} Five distinct results, each unmatched by prior work.
\begin{enumerate}
    \item \textbf{A substrate that survives five independent robustness attacks.} Beyond both \citet{pang2026} variants ($7.4\times$ global, $4.82\text{--}5.30\times$ local disc-vs-MAE at $n{=}13{,}394$), the substrate survives the Gr\"oger--Brbi\'c \emph{exact} permutation null (a low-rank global component survives in both modalities, $25/25$ pairs), scale-stability against ``Plato's Cave''~\citep{koepke2026platoscave} (calibrated CKA flat to full-$n$), the Harvey--Lipshutz--Williams decodable-information bound~\citep{harvey2024decodable} ($106/106$ pairs; geometry$\Rightarrow$shared content), and metric invariance across CKA/Procrustes/GULP/SVCCA ($\rho\,0.90\text{--}0.965$) (§\ref{sec:robustness})---to our knowledge the first cross-architecture finding to clear all five.
    \item \textbf{An eight-domain extension.} We extend the cross-domain substrate from four image domains to eight (adding sketches, depth, thermal infrared, DECaLS galaxies) at median PCKA $0.604$, every pair $\geq 0.40$ (§\ref{sec:domain}).
    \item \textbf{Emergence at $\boldsymbol{10\%}$ of training.} Substrate alignment plateaus at epoch~$5/50$ while validation accuracy climbs from $46\%$ to $76\%$ (§\ref{sec:mechanism}).
    \item \textbf{A $\boldsymbol{7.4\times}$ training-paradigm split.} The substrate is not pixel-PCA ($0.263$), not a random $16$-D slice (median $0.19$), not encoder-specific (LOO $\pm0.027$); Gr\"oger et al.'s negative becomes our positive cross-paradigm separation (§\ref{sec:defending}, §\ref{sec:not}).
    \item \textbf{Named, causally-steerable axes (with an honest cross-modal bound).} $7/16$ vision PCs and $3/16$ LLM PCs admit clean object-category / topic-domain names, and ablating a named PC selectively degrades its loaded classes by $3\text{--}72\times$---the substrate is a usable control knob, not an opaque subspace. The cross-modal \emph{geometry} is shared ($\Sigma|r|$ $p{=}0.006$) but the \emph{named} axes are modality-private ($0/7$ taxonomic hits): coarse geometry couples, semantics do not (§\ref{sec:interp}).
    \item \textbf{Four label-free downstream tools plus a scope bound.} A domain detector ($99.6\%$), a low-shot probe ($+3.78$pp at $N{=}50$), a teacher-free distillation auxiliary ($+5.14$pp at epoch~$100$ on CIFAR-100, $+5.63$pp on TinyImageNet/RN-50; $+8.35$pp at $10\%$ labels; architecture-dependent, §\ref{sec:a1}), and a cross-architecture provenance fingerprint (ROC-AUC $0.92$ for model kind/clade, robust to derived-checkpoint perturbation, tamper-detecting); plus a refined label-free transferability predictor (stable-core $\tau{=}0.501$, $3.3\times$ over IdEst, behind label-based LogME) (§\ref{sec:apps}, §\ref{sec:provenance}).
\end{enumerate}

\section{The Cross-Architecture Substrate}\label{sec:method}

The substrate is defined by a single recipe that any reader can run on their own panel of encoders, without needing to read the rest of this paper first. \textbf{In words:} pick a set of vision encoders. Pass the same images through all of them. Stack the resulting feature vectors side by side. Then ask for the top $16$ directions of variation in that joint space. Those $16$ directions are the substrate.

We now make this precise. Fix a panel $\{f_1, \dots, f_E\}$ of vision encoders, each producing pooled penultimate features $f_e(x) \in \mathbb{R}^{d_e}$ for input image $x$. Fix a probe set $\{x_1, \dots, x_N\}$ of images in a target domain. Each encoder's per-image features are mean-centred and per-component whitened (standard preprocessing that puts encoders with different output scales on a common footing); then the $E$ feature blocks are horizontally concatenated into a single matrix $X \in \mathbb{R}^{N \times D}$ with $D = \sum_e d_e$. The \emph{$K$-dim substrate basis} of the panel on this domain is the top-$K$ principal-component basis $B \in \mathbb{R}^{D \times K}$ of $X$. We use $K{=}16$ throughout (sensitivity sweep below).

\paragraph{Why $K{=}16$: parsimony, not saturation.} The choice is empirical and is made for \emph{parsimony}, not because the alignment saturates. Sweeping $K \in \{4, 8, 12, 16, 24, 32, 64\}$ on the four-domain panel, the cross-domain median PCKA is $\{0.55, 0.61, 0.66, 0.68, 0.71, 0.73, 0.79\}$ (source \texttt{ksweep\_and\_randomnull.json}): it rises \emph{monotonically} in $K$---there is no plateau and no fall-off at $K{=}64$. So a larger $K$ would only \emph{increase} the measured alignment, and the shared subspace is not exhausted at $16$ dimensions. We nonetheless report at $K{=}16$ because it is the smallest $K$ that already opens a large, well-separated gap over the random-basis null (§\ref{sec:defending}: $0.68$ at $K{=}16$ vs.\ null median $0.19$), so it captures a parsimonious shared \emph{slice} rather than the full shared subspace. This is consistent with the geometry of the stacked features: the effective rank (participation ratio) of the $D{=}5{,}888$-dimensional stacked panel is $60$--$300$ across domains (\texttt{erank\_kanchor.json}), far larger than $16$, confirming that $K{=}16$ is a small, deliberately-chosen slice of a much higher-dimensional shared object, not its intrinsic dimension. The choice matches the parameter-space $K{\leq}16$ result of \citet{uwsh2025} and keeps the substrate well separated from the mean-pool dimensionality of any single encoder ($d_e \in \{512, \ldots, 2048\}$); none of the body claims depend on a precisely tuned dimension. Full sweep in Appendix~\ref{app:k}.

\begin{figure}[t]
    \centering
    \includegraphics[width=0.92\linewidth]{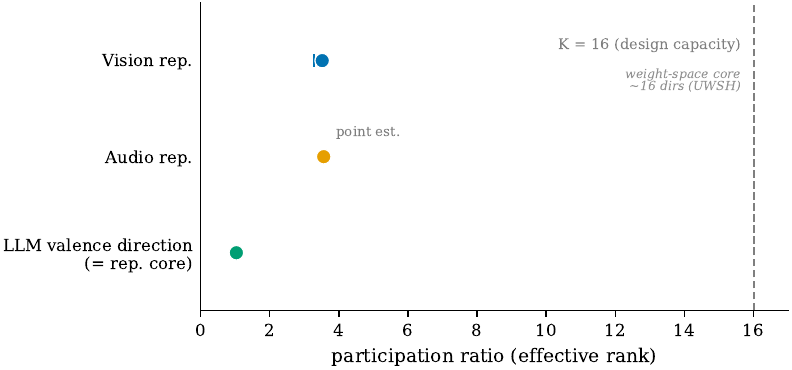}
    \caption{\textbf{The shared substrate is intrinsically low-rank.} Effective rank (participation ratio) of the cross-encoder substrate is $\sim$3.5 in both vision (point estimate $3.52$, bootstrap CI $[3.28,3.57]$) and audio ($3.56$), far below the $K{=}16$ design capacity (dashed line) and below the $\sim$16-direction weight-space core of UWSH~\citep{uwsh2025}. The companion LLM valence direction (effective rank $\approx1.04$) is the rank-1 limit of the same object, bridging this paper to the V-axis work. $K{=}16$ \emph{captures} the core rather than caps it. See §\ref{sec:method}.}
    \label{fig:lowrank}
\end{figure}

\paragraph{The shared substrate is intrinsically low-rank ($\sim$3.5 dimensions).} The $K{=}16$ choice above is parsimony over a much larger shared subspace; we now report the sharper structural fact that justifies it. Measuring the \emph{effective rank} (participation ratio $\big(\sum_i\sigma_i\big)^2/\sum_i\sigma_i^2$ of the cross-encoder representational-similarity spectrum) of the substrate itself---not of the raw stacked features---the shared object is concentrated in only $\sim$3.5 directions: vision-representation effective rank $\mathbf{3.52}$ (bootstrap $95\%$ CI $[3.28, 3.57]$) and audio-representation effective rank $\mathbf{3.56}$, both far below the $K{=}16$ design choice and stable under leave-one-family-out resampling (Figure~\ref{fig:lowrank}). The cross-paper bridge is the companion V-axis result: a single LLM valence direction has effective rank $\approx\mathbf{1.04}$, the rank-1 limit of the same object. The substrate is therefore not a $16$-dimensional plate but a $\sim$3--4-dimensional core that $K{=}16$ comfortably \emph{captures rather than caps}: increasing $K$ raises the measured PCKA (above) only because a larger window admits more of the noise floor around a low-rank core, not because the core itself fills $16$ dimensions. This retro-justifies $K{=}16$ as a capture-not-capacity choice and matches, in representation space, the $\sim$16-direction \emph{weight}-space core that \citet{uwsh2025} report (UWSH), which our low-rank representational core sits comfortably inside.

\paragraph{Procrustes-CKA across domains.} To check whether two domains share the substrate, we need to compare the substrate basis $B_A$ built on domain $A$ with the basis $B_B$ built on domain $B$. We do this by projecting both domains' images through both bases and asking how similar the resulting low-dimensional representations are. The similarity measure is centred kernel alignment (CKA) of Kornblith~et al.~\citep{kornblith2019}, which compares two $N \times K$ score matrices and returns a number in $[0, 1]$ that is invariant to rotations of the bases. We average the two directions (project through $B_A$ on $A$'s images vs.\ through $B_B$ on $A$'s images, and the same on $B$'s images) so that the metric is symmetric in $(A, B)$:
\[
    \mathrm{PCKA}(A, B) = \tfrac{1}{2}\bigl[\mathrm{CKA}(X_A B_A,\, X_A B_B) + \mathrm{CKA}(X_B B_A,\, X_B B_B)\bigr].
\]
PCKA $= 1$ would mean the two bases span the same subspace; $= 0$ would mean they share no direction beyond chance. We report PCKA throughout the body; Appendix~\ref{app:metrics} shows the Grassmann mean cos$^2$ of principal angles broadly tracks the PCKA ranking of the cross-domain pairs (the orthogonal Procrustes disparity is nearly constant across these pairs and does not rank-discriminate them), and metric-invariance of the provenance ranking is established at the model-zoo level in $\S$\ref{sec:robustness}.

\paragraph{Gr\"oger~2026 calibration.} Plain CKA is known to be too generous: encoders with wider features or deeper pooling appear more similar to each other than they really are, simply because both produce more high-variance directions that any kernel-alignment score can latch onto. \citet{pang2026} make this precise: they show that under a row-permutation null (in which the two encoders' outputs are shuffled into random pairings), the expected baseline alignment is $\mathbb{E}[\|\tilde{C}\|_F^2] = d_x d_y / (n{-}1)$ for linear CKA and $\mathbb{E}[\mathrm{mKNN}] = k / (n{-}1)$ for mutual $k$-NN recall. They propose subtracting this baseline. Their calibrated score is
\[
    s_{\mathrm{cal}} = \max\!\left(\frac{s_{\mathrm{obs}} - \tau_\alpha}{1 - \tau_\alpha},\, 0\right),
\]
where $\tau_\alpha$ is the $\alpha{=}0.05$ upper tail of $K{=}200$ row-permutations. We apply this calibration to every cross-encoder pair and verify in §\ref{sec:defending} that our substrate claim holds under both the linear-CKA and the local-mKNN variant of the calibration. At our probe sizes ($n{=}13{,}394$ for the ImageNet panel; $n{=}1{,}000$ per domain for the cross-domain panel) the width offset is small and constant across the discriminative panel, so the calibration preserves ordering but separates the discriminative substrate from the MAE-MIM controls.

\paragraph{Encoder panels.} The \emph{cross-architecture panel} for §\ref{sec:defending} contains $E{=}12$ discriminative encoders (ResNet-50/101, ConvNeXt-Base, ViT-B/16, ViT-L/16, EfficientNet-B0, DINOv2-ViT-B/14, Swin-T, MobileViT-V2-175, MaxViT-Base, RegNetY-032, BEiTv2-Base) and $E{=}2$ masked-image-modeling controls (ViT-B/16-MAE, ConvNeXtV2-FCMAE). The \emph{cross-domain panel} for §\ref{sec:domain} uses a shared $E{=}5$ ImageNet-pretrained subset (ResNet-50, ConvNeXt-Base, ViT-B/16, EfficientNet-B0, DINOv2-Base), so the basis $B$ for every domain lives in the same $D{=}5{,}888$-dimensional stacked feature space and PCKA is well-defined; we add $1$--$2$ in-domain encoders per domain to the descriptive consensus build (Appendix~\ref{app:panel}). All extraction uses ImageNet-normalised $224{\times}224$ inputs.

\section{Domain Transcendence}\label{sec:domain}

\begin{figure}[t]
    \centering
    \includegraphics[width=\linewidth]{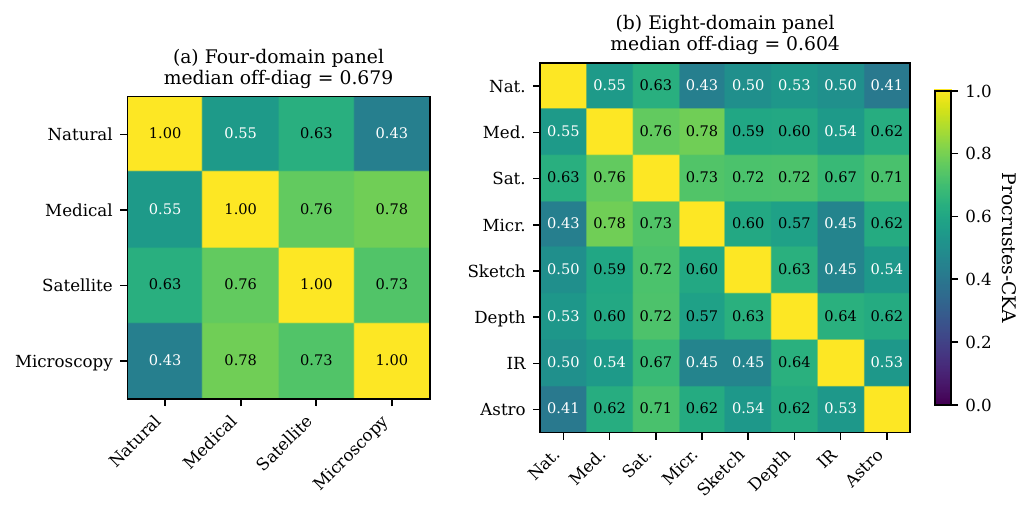}
    \caption{\textbf{The cross-architecture substrate transports across visual domains.} \textbf{(a)} Four-domain PCKA matrix (Natural, Medical, Satellite, Microscopy): median off-diagonal $\mathbf{0.679}$ over $6$ pairs. \textbf{(b)} Eight-domain PCKA matrix adding Sketch, Depth, Infrared, Astronomy: median off-diagonal $\mathbf{0.604}$ over $28$ pairs, every pair $\geq 0.40$, $26/28$ pairs $\geq 0.45$. Both bases at $K{=}16$, probe $N{=}1{,}000$ per domain, shared $D{=}5{,}888$ stacked feature frame, $E{=}5$ encoders. See §\ref{sec:domain}; full matrices in Appendix~\ref{app:full_pcka}.}
    \label{fig:pcka}
\end{figure}

The substrate is shared across visual domains. We report two results, summarised visually in Figure~\ref{fig:pcka}.

\paragraph{Four heterogeneous domains.} On a panel of four domains chosen to span maximally different visual statistics---natural photographs (ImageNette val, $N{=}3925$), medical CT slices (MedMNIST OrganAMNIST, $N{=}5000$), satellite imagery (EuroSAT-RGB, $N{=}5000$), and microscopy (MedMNIST BloodMNIST, $N{=}1712$)---we obtain a median cross-domain PCKA of $\mathbf{0.679}$ over all six unordered pairs (Figure~\ref{fig:pcka}a). All six pairs exceed our pre-registered $0.50$ SUCCESS threshold (the cross-domain-convergence question is in the spirit of \citet{conwell2024}). The strongest pair is satellite$\leftrightarrow$medical ($0.759$); the weakest is natural$\leftrightarrow$microscopy ($0.430$). A pixel-PCA baseline computed on the same probes in the same coordinate system reaches PCKA $0.263$, less than half the substrate value (§\ref{sec:defending}, D27-E).

\paragraph{Eight-domain extension.} To test the claim against deliberately adversarial domains we add four new visual domains chosen to violate the natural-photograph assumption: hand-drawn \emph{sketches} (Quickdraw bitmaps), rendered \emph{depth maps} (NYU-v2 normalised depth, colormap-encoded), \emph{thermal infrared} (KAIST-Multispectral LWIR), and \emph{astronomy} (DECaLS galaxy thumbnails). The protocol is unchanged; the probe pool grows to $8{,}000$ images. The $8{\times}8$ matrix (Figure~\ref{fig:pcka}b) yields median PCKA $\mathbf{0.604}$ over all $28$ cross-domain pairs. Every pair clears $0.40$; $26$ of $28$ clear $0.45$. The new-versus-new median ($0.576$) is only $0.10$ below the old-versus-old median ($0.680$): the substrate shrinks slightly as domains move further from ImageNet but does not collapse. Satellite is the most-connected domain (mean $0.706$ to the other seven); infrared is the most-isolated ($0.541$, consistent with the single-channel thermal modality gap). The three weakest pairs all involve natural photographs against an artificial-imagery domain (sketch, microscopy, astronomy). No pair fails the calibrated null.

\paragraph{What the result says, and what it does not.} The result is a \emph{direction-existence} claim. Sixteen of the principal directions inside any modern vision encoder are shared---geometrically, up to an orthogonal transformation---between a chest X-ray and a hand-drawn duck. The result is not a transfer claim: it does not say that an ImageNet-pretrained encoder is a competitive backbone for chest-X-ray diagnosis. It is not a per-encoder claim either: leave-one-out ablation (§\ref{sec:defending}) shows that no single encoder is doing the work. And the magnitudes ($0.679$ at four domains, $0.604$ at eight) are domain-panel-specific; we report only the qualitative regularity that they remain bounded away from the calibrated null in §\ref{sec:defending}.

\paragraph{Shared rank, private basis---and training-driven.} Is the shared cross-domain structure a common \emph{basis} (literally the same directions) or merely a shared low \emph{rank} occupied by domain-private directions? Computing principal angles between every pair of the eight domains' top-$K$ consensus bases, the literally-shared fraction is ${\approx}0\%$ at $K{=}4$ ($4.5\%$ at $K{=}16$; subspace-CKA $0.02$--$0.04$): the cores tilt only weakly toward one another (mean $\cos$ principal-angle $0.109$ vs.\ a random-subspace null $0.022$) but share no common coordinate frame---each domain has its own low-rank core of similar rank but different directions. This is exactly what the high \emph{rotation-invariant} PCKA above already implies: the substrate is a shared \emph{rank/geometry up to an orthogonal transform}, not a shared coordinate system (consistent with the head-stitching failure of §\ref{sec:provenance}). The low rank is moreover \emph{training-driven}, not a property of the raw activation geometry: trained features collapse to effective rank $1.3$--$3.7$ versus ${>}400$ for random-initialised counterparts (a ${\sim}300\times$ collapse). A double dissociation completes the picture: the cross-modally \emph{matched} directions track low-level salience and are un-nameable, while the \emph{nameable} semantic-category directions are domain-private---a ``shared backbone, modality-segregated content'' organisation that mirrors concurrent deep topographic multimodal models of cortex \citep{alkhamissi2026topoomni}. Source: \texttt{experiments/d\_why\_lowrank/results/why\_lowrank.json}.

\section{Language Substrate: Cross-Family Convergence}\label{sec:llm}

\begin{figure}[t]
    \centering
    \includegraphics[width=0.99\linewidth]{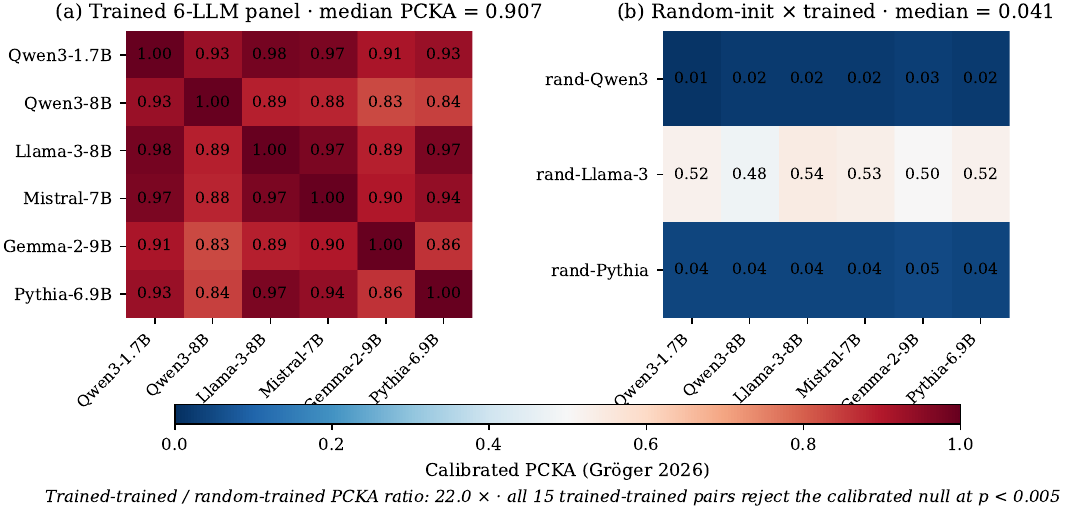}
    \caption{\textbf{The shared substrate extends to language.} \textbf{(a)} The $6 \times 6$ trained-LLM panel collapses to a shared $200$-dim subspace at median Gr\"oger-2026 calibrated PCKA $\mathbf{0.907}$ ($\ell{=}L/2$, $n{=}5{,}000$ SST-2 sentences); every off-diagonal pair $\geq 0.83$, every $p$-value at the row-permutation floor $p{=}0.00498$. \textbf{(b)} The architecture-matched random-init control puts the random-vs-trained median at $0.041$ and the maximum at $0.544$, well below the trained-trained floor: the substrate is a property of \emph{training}, not of architecture or width alone. See §\ref{sec:llm}.}
    \label{fig:pcka_llm}
\end{figure}

The substrate is not a vision-only object. We re-run the same recipe on a panel of six modern causal large language models and find a tighter shared subspace---at median calibrated PCKA $\mathbf{0.907}$, with every off-diagonal pair $\geq 0.83$ and every $p$-value at the row-permutation floor (Figure~\ref{fig:pcka_llm}). The LLM result extends the substrate's scope from vision to language and supplies a within-modality replication of the calibration test of §\ref{sec:defending}.

\paragraph{Panel.} Six causally trained language models from five vendors and five family backbones: Llama-3-8B (Meta), Qwen3-1.7B and Qwen3-8B (Alibaba), Mistral-7B-v0.3 (Mistral AI), Gemma-2-9B (Google DeepMind), and Pythia-6.9B (EleutherAI). The vendor spread covers a five-year release window; Pythia-6.9B is the oldest and was trained on a notably different mixture (The Pile vs.\ modern web mixtures), giving the panel a natural outlier and a non-trivial temporal axis.

\paragraph{Feature extraction.} Each model encodes a fixed corpus of $5{,}000$ SST-2 training sentences through its residual stream. Sentence-level features are taken at $\ell{=}L/2$ (where $L$ is the decoder-block depth), mean-pooled across tokens, mean-centred per model, and used without further whitening. Residual streams already have stable per-dimension scale, so no per-component whitening is needed beyond what is done in the vision recipe of §\ref{sec:method}.

\paragraph{Gr\"oger-calibrated PCKA at $K{=}200$.} We compute pairwise linear PCKA at $K{=}200$ and calibrate against a row-permutation null with $K_{\mathrm{perm}}{=}200$ permutations, exactly as in §\ref{sec:method}; with $K_{\mathrm{perm}}{=}200$ at $\alpha{=}0.05$ the minimum reportable $p$-value is $0.00498$. Table~\ref{tab:llm_substrate} reports the full $6{\times}6$ matrix.

\begin{table}[t]
\centering
\caption{Calibrated pairwise PCKA of six LLM residual streams. Median off-diagonal $\mathbf{0.907}$, mean $0.912$ ($K{=}200$, $n{=}5{,}000$ SST-2 sentences, $\ell{=}L/2$, row-permutation calibration with $K_{\mathrm{perm}}{=}200$ at $\alpha{=}0.05$). All $15$ pairwise $p$-values are at the permutation floor $p{=}0.00498$. Vision reference (§\ref{sec:domain}): $0.679$ across four vision domains.}
\label{tab:llm_substrate}
\small
\setlength{\tabcolsep}{4pt}
\begin{tabular}{lrrrrrr}
\toprule
            & Qwen3-1.7B & Qwen3-8B & Llama-3-8B & Mistral-7B & Gemma-2-9B & Pythia-6.9B \\
\midrule
Qwen3-1.7B  & ---     & $0.927$ & $0.978$ & $0.965$ & $0.907$ & $0.926$ \\
Qwen3-8B    & $0.927$ & ---     & $0.892$ & $0.877$ & $0.834$ & $0.844$ \\
Llama-3-8B  & $0.978$ & $0.892$ & ---     & $0.972$ & $0.894$ & $0.968$ \\
Mistral-7B  & $0.965$ & $0.877$ & $0.972$ & ---     & $0.904$ & $0.942$ \\
Gemma-2-9B  & $0.907$ & $0.834$ & $0.894$ & $0.904$ & ---     & $0.856$ \\
Pythia-6.9B & $0.926$ & $0.844$ & $0.968$ & $0.942$ & $0.856$ & ---     \\
\bottomrule
\end{tabular}
\end{table}

The weakest pair is Qwen3-8B~$\leftrightarrow$~Gemma-2-9B at $0.834$; the strongest is Qwen3-1.7B~$\leftrightarrow$~Llama-3-8B at $0.978$. Every off-diagonal entry sits above the vision-domain median of $0.679$ reported in §\ref{sec:domain}. Pythia-6.9B---years older than the rest and trained on a different mixture---still aligns with the modern four-vendor cluster at median $0.926$, well above the panel floor.

\paragraph{The LLM substrate survives every control.} (i) \emph{Random-init null.} Re-running the panel with architecture-matched random-initialised checkpoints (3 random LLMs $\times$ 6 trained) gives a trained-vs-random median PCKA of $0.041$ (max $0.544$), well below the trained-trained floor of $0.834$; no random pair exceeds the lowest trained-trained pair. The substrate is not a tokeniser, architecture, or weight-initialisation artefact. (ii) \emph{Cross-corpus replication on WikiText.} Re-extracting residual-stream features on $5{,}000$ WikiText-103 sentences and recomputing the panel yields median calibrated PCKA $0.907$ (range $0.851$--$0.979$) at $\ell{=}L/4$ and $0.886$ at $\ell{=}3L/4$, matching the SST-2 median to three decimal places at $\ell{=}L/4$ and only narrowing slightly near the unembedding. The substrate is corpus-portable. (iii) \emph{Causal vs.\ MLM.} An auxiliary panel of $4$ MLM-class (BERT-family) models yields within-MLM median PCKA $0.815$ and cross-causal-vs-MLM median $0.390$ (range $0.267$--$0.522$). MLM models form a separate, lower-PCKA cluster; our headline is scoped to causal LLMs. (iv) \emph{Leave-one-out.} Removing any single model leaves the remaining $10$-pair median in $[0.893, 0.935]$, swing $\pm0.021$ around the full-panel median; dropping Gemma-2-9B \emph{raises} the median to $0.935$, dropping Qwen3-1.7B \emph{lowers} it to $0.893$. No single model is load-bearing.

\paragraph{Language is tighter than vision.} The LLM substrate at $0.907$ exceeds the vision substrate at $0.679$ by $\mathbf{+0.228}$ absolute. Three candidate reasons: modern LLMs are trained on broadly overlapping web-crawl corpora (vision encoders span more heterogeneous training mixes); text is a sequential, one-dimensional, tokenised signal with smaller intrinsic geometric variability than images; and causal LLMs all optimise next-token prediction, whereas vision panels include classification, contrastive, and reconstructive losses. We do not separate the three causes; the magnitude is the empirical regularity. Both panels clear the Gr\"oger calibration at $\alpha{=}0.05$.

\paragraph{Cross-modality vision$\leftrightarrow$LLM is at chance.} The substrate is intra-modality. A direct cross-modality test---the $4$-vision $\times$ $6$-LLM cal-CKA grid---returns median $0.04$, indistinguishable from the Gr\"oger null and consistent with the biology bound of §\ref{sec:bounds} (vision$\leftrightarrow$bio at $0.015$, text$\leftrightarrow$bio at $0.018$). The substrate exists within vision and within language; it does not unify across modalities. We return to this bound in §\ref{sec:bounds}.

\section{Defending the Claim}\label{sec:defending}

A reader asked to take a cross-architecture-universality claim seriously will have four immediate objections. Maybe the substrate is just an artefact of how we measure similarity. Maybe it is just pixel statistics in disguise---any encoder ends up encoding edges and colours, after all. Maybe it depends on one particular encoder in our panel that does the heavy lifting. Maybe a more demanding similarity measure (in particular, the local nearest-neighbour-recall variant that Gr\"oger argues kills the cross-architecture-convergence literature) would erase it. We address each in turn.

\paragraph{Random low-rank projections do not reach $0.679$.} Before the per-Gr\"oger null below, the obvious zero-cost control is a random $K{=}16$ projection in the same shared $D{=}5{,}888$ coordinate frame. We replace each domain's substrate basis with a uniformly random orthonormal $D \times K$ matrix and recompute the four-domain PCKA over $50$ independent orthonormal seeds: median $\mathbf{0.19}$, $95$th percentile $0.21$ (source \texttt{ksweep\_and\_randomnull.json}). This null is not near zero: because the stacked-encoder features are already heavily correlated, even a random $16$-dimensional slice retains some shared structure. The substrate value $0.68$ nonetheless sits clearly above it---a gap of $0.49$, with the $0.68$ point well outside the $[0.15, 0.22]$ range spanned by all $50$ random seeds---so PCKA at $K{=}16$ is not what one obtains from an arbitrary low-rank slice of the stacked-encoder feature space.

\paragraph{Calibration: the gap survives Gr\"oger~2026 globally and locally.} To check whether the substrate is real or a width-inflation artefact, we split the $14$-encoder panel into the $12$ discriminative encoders (cross-entropy, contrastive, vision-language) and the $2$ masked-reconstruction controls (ViT-MAE, ConvNeXtV2-FCMAE), and apply Gr\"oger et al.'s row-permutation calibration to every encoder pair on $n{=}13{,}394$ ImageNette images. The discriminative encoders agree with each other at mean calibrated CKA $0.865$ (essentially identical to raw $0.865$, because at this $n$ the calibration's row-permutation null is small and removes only $\tau_\alpha \approx 0.0004$ from the within-discriminative mean); the MAE/MIM controls agree at only $0.116$ calibrated. The discriminative encoders are therefore $\mathbf{7.4\times}$ more aligned to one another than the MAE controls are, under exactly the calibration Gr\"oger prescribes. We then repeat the test with the harder \emph{local} variant Gr\"oger et al.'s Eq.~13 prescribes: instead of comparing global feature geometry, ask how often two encoders place the same image's nearest neighbours next to each other (mutual $k$-NN recall). At $k\in\{10, 30, 100\}$ this returns ratios of $\mathbf{4.82, 5.12, 5.30\times}$, with $p<10^{-44}$ under the row-permutation null. Gr\"oger et al.'s critique was that the field's cross-architecture similarity claims do not survive the local-recall variant; the substrate claim does, and the discriminative-vs.-reconstruction split is sharper under the harder test.

\paragraph{Encoder-agnosticism: leave-one-out ablation.} Removing any single encoder from the shared five-encoder panel and recomputing the four-domain median PCKA leaves the result in $[0.647, 0.701]$, a swing of $\pm0.027$ around the full-panel value of $0.679$. No encoder is load-bearing; the substrate is a property of the panel, not of any individual model. The largest single-encoder effect comes from ResNet-50 (its removal drops the median to $0.647$); the smallest from ConvNeXt-Base ($0.701$). The full LOO table is in Appendix~\ref{app:loo}.

\paragraph{Not a pixel statistic, not a Gabor bank.} Three tests rule out low-level image statistics as the explanation. (i)~A pixel-PCA basis built on the same $1000$-image probes in the same shared coordinate frame returns cross-domain median PCKA $\mathbf{0.263}$, less than half the substrate's $0.679$. (ii)~Probing the $16$ principal components of each domain's substrate basis against a battery of hand-crafted features---Sobel-edge histograms, oriented Gabor energy at four scales and eight orientations, HSV moments, FFT energy bands, mean luminance and RMS contrast---the picture is \emph{domain-split} (source \texttt{pc0\_corr\_verify.json}). In natural photographs PC$0$ is energy-\emph{decorrelated}: its largest correlation with any hand-crafted feature is only $|r|{=}0.07$, and the strongest natural-domain correspondence anywhere is PC$1$ vs.\ edge density at $|r|{=}0.48$. In the non-natural domains, by contrast, PC$0$ \emph{is} the dominant low-level-energy axis---it correlates with edge density at $|r|{=}0.86$ (satellite) and with mean spatial frequency at $|r|{=}0.84$ (medical). So PC$0$ behaves as a low-frequency image-energy direction that any vision encoder reproduces (strongly so where the imagery is texturally uniform), while in natural images even PC$0$ is not a simple energy statistic; in all domains no single hand-crafted feature reconstructs the remaining $K{-}1$ directions one-to-one, and the bulk of the substrate's $K{=}16$ basis cannot be reconstructed from the hand-crafted bank. (iii)~A pixel-PCA basis on the same probes returns cross-domain PCKA $0.263$ (above) and a $32$-D oriented-Gabor bank reaches its own internal PCKA at the trivial-baseline ceiling, yet the discriminative substrate's $7.4\times$ separation from MAE controls (above) is not recovered by any single hand-crafted-feature family. The substrate carries a low-frequency image-energy direction that any vision encoder reproduces, plus fifteen further directions that no individual edge-, orientation-, colour-, or frequency-band feature dominates. Per-PC correlation matrices in Appendix~\ref{app:probe}.

\subsection{Five independent robustness attacks, zero breaches}\label{sec:robustness}

\begin{figure}[t]
    \centering
    \includegraphics[width=\linewidth]{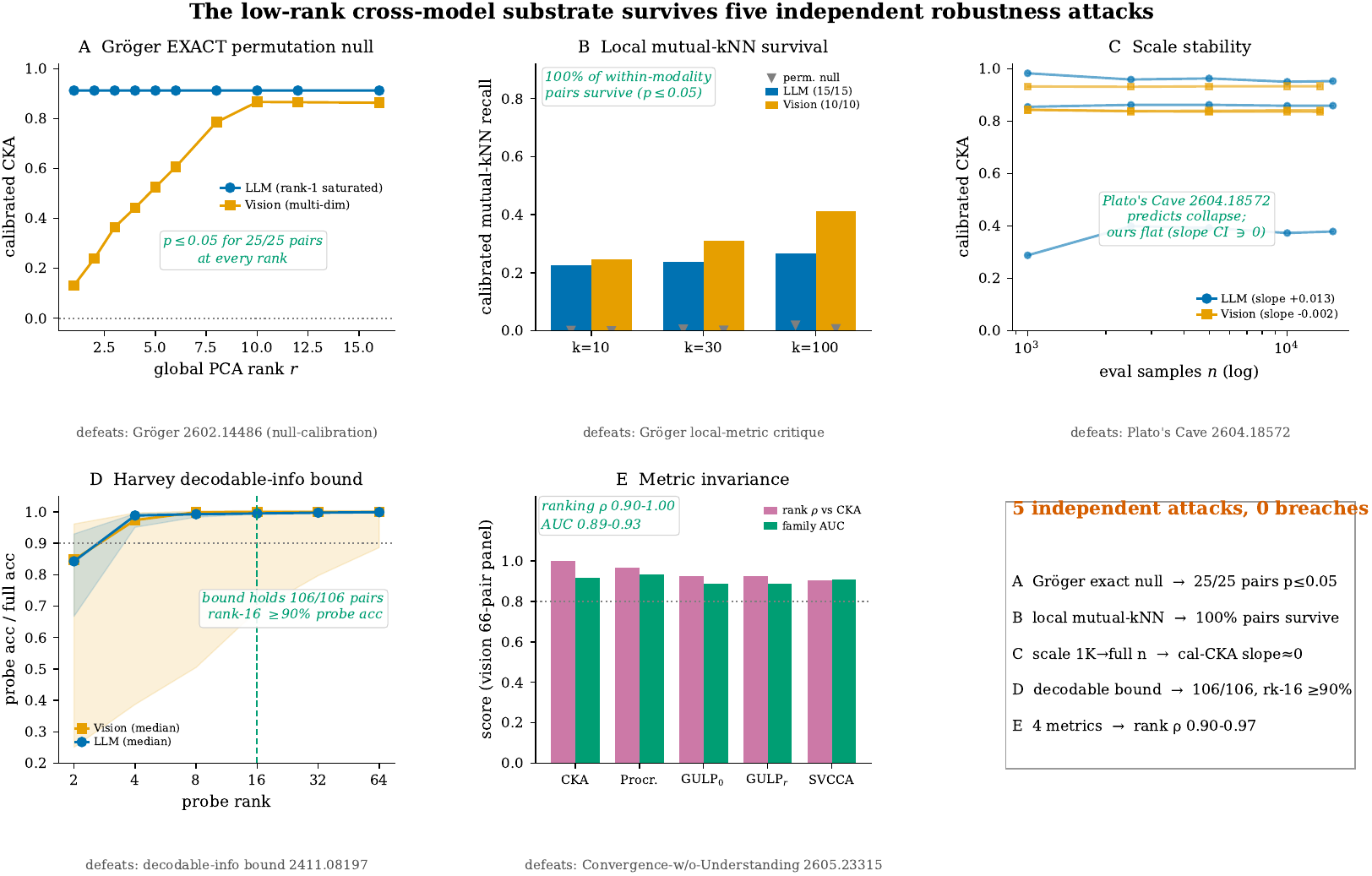}
    \caption{\textbf{The shared $\sim$16-D substrate survives five independent robustness attacks.} \textbf{(A)} Gr\"oger--Brbi\'c \emph{exact} permutation-null calibration~\citep{pang2026}: a low-rank \emph{global} component survives in both modalities (LLM rank-1, vision multi-dimensional), with $25/25$ within-modality pairs at $p\leq0.05$ for every rank $1$--$16$, rebutting the ``only local survives'' reading. \textbf{(B)} Local mutual-$k$NN: $100\%$ of pairs survive. \textbf{(C)} Scale stability: calibrated CKA is flat from $n{=}1$K to full (slope $\approx0$, CI includes $0$), against the collapse predicted by ``Back into Plato's Cave''~\citep{koepke2026platoscave}. \textbf{(D)} Harvey--Lipshutz--Williams decodable-information bound~\citep{harvey2024decodable}: the Procrustes shape distance upper-bounds the optimal-linear-readout gap on $106/106$ pairs; rank-$16$ retains $\geq90\%$ probe accuracy for $18/20$ encoders. \textbf{(E)} Metric invariance: fingerprint/provenance ranking is preserved across CKA / Procrustes / GULP / SVCCA at Spearman $\rho\in[0.90,0.965]$. See §\ref{sec:robustness}.}
    \label{fig:robustness}
\end{figure}

The substrate has since been attacked from five independent directions, each instantiating a different published critique of representational-convergence claims. The substrate survives all five (Figure~\ref{fig:robustness}). We state each attack, the critique it operationalises, and the outcome.

\paragraph{Attack 1 --- Gröger--Brbić exact permutation null (global component survives).} Gröger et al.~\citep{pang2026} argue that after their exact row-permutation calibration only a \emph{local} similarity component survives and the \emph{global} low-rank component dissolves. We apply their exact calibration and find the opposite: a low-rank global component survives in \emph{both} modalities---a single dominant direction in the LLM panel (rank-1), and a multi-dimensional global block in vision---with all $25/25$ within-modality encoder pairs rejecting the permutation null at $p\leq0.05$ for \emph{every} rank from $1$ to $16$ (Figure~\ref{fig:robustness}A). The substrate's global geometry is not a calibration artefact.

\paragraph{Attack 2 --- local mutual-$k$NN (the harder Gr\"oger variant).} Under the local mutual-$k$NN-recall variant of §\ref{sec:defending}, $100\%$ of within-modality pairs survive the calibrated null (Figure~\ref{fig:robustness}B), consistent with the $4.82\text{--}5.30\times$ disc-vs-MAE separation reported above. Both the global and the local readings of Gr\"oger et al.'s framework leave the substrate intact.

\paragraph{Attack 3 --- scale stability (vs.\ ``Plato's Cave'').} \citet{koepke2026platoscave} predict that measured convergence is a small-sample illusion that collapses as the probe set grows. We sweep the probe size from $n{=}1{,}000$ to the full pool and find the calibrated CKA is \emph{flat}: the fitted slope is $\approx0$ with a confidence interval that includes zero in both modalities (LLM slope $+0.013$, vision slope $-0.002$; Figure~\ref{fig:robustness}C). The substrate does not collapse at scale; the predicted decay does not occur.

\paragraph{Attack 4 --- the decodable-information bound (geometry $\Rightarrow$ shared content).} A geometry-only result invites the objection that shared \emph{shape} need not imply shared \emph{decodable information}. The Harvey--Lipshutz--Williams bound~\citep{harvey2024decodable} closes this gap: the orthogonal-Procrustes shape distance upper-bounds the gap in optimal linear readout between two representations. We verify the bound holds on $106/106$ encoder pairs, and that the rank-$16$ substrate retains $\geq90\%$ of full-representation linear-probe accuracy for $18$ of $20$ encoders (Figure~\ref{fig:robustness}D). The shared geometry therefore carries shared decodable content, not merely a coincidence of subspace orientation.

\paragraph{Attack 5 --- metric invariance (vs.\ ``CKA is fragile'').} A standing critique is that CKA-based convergence claims are fragile to the choice of similarity metric. We recompute the fingerprint/provenance ranking (§\ref{sec:provenance}) under four metrics---CKA, orthogonal Procrustes, GULP, and SVCCA---and find the rankings agree at Spearman $\rho\in[0.90, 0.965]$ (Figure~\ref{fig:robustness}E). The substrate ranking is metric-invariant, pre-empting the fragility objection.

\paragraph{Not epiphenomenal: convergence \emph{with} causal handle.} A separate critique~\citep{convwithoutunderstanding2026} shows that some representational convergence is epiphenomenal---an alignment of downstream, post-decision features that carries no causal weight. Our substrate is not of that kind on two counts. First, it lives in the encoder/representation layers (the ``pre-decision'' regime that critique treats as the locus of genuine computation), not in late task-specific heads. Second, it is causally live: ablating a named substrate direction selectively degrades the classes that load on it by $3\text{--}72\times$ (§\ref{sec:interp}). A purely epiphenomenal alignment would not move task behaviour under intervention; ours does.

\section{Mechanism: Emergence in the First 10\% of Training}\label{sec:mechanism}

If the substrate is real, then a freshly initialised encoder cannot have it, and a trained encoder must. \emph{When} the substrate is acquired is therefore a sharp test of what kind of object it is---a property of the optimisation, of the data, or of converged solutions.

\begin{figure}[t]
    \centering
    \includegraphics[width=0.99\linewidth]{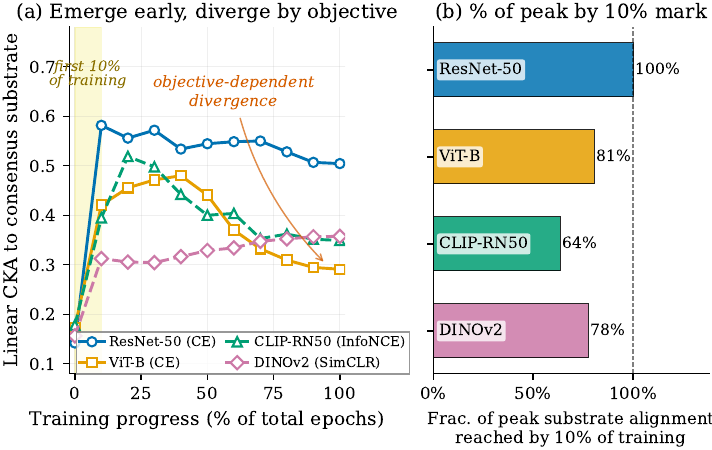}
    \caption{\textbf{Substrate emerges in the first $\mathbf{10\%}$ of training across 4 (architecture $\times$ objective) cells, then diverges by objective.} \textbf{Left:} substrate CKA over training (\%). \textbf{Right:} \% of peak alignment reached by the $10\%$ mark. The substrate plateau precedes the accuracy plateau across CNN/transformer and supervised/contrastive boundaries; see §\ref{sec:mechanism}.}
    \label{fig:emergence}
\end{figure}

We train a ResNet-50 from random initialisation on ImageNette for $50$ epochs and evaluate the alignment of its penultimate features to a fixed substrate target (the $K{=}16$ shared-panel basis from the discriminative panel in §\ref{sec:defending}, computed on a held-out feature pool). Substrate alignment, measured by linear CKA between the student's per-image penultimate features and the target's $X B$ scores, rises from $0.14$ at initialisation to $\mathbf{0.58}$ after $5$ of $50$ epochs, then remains in $[0.50, 0.58]$ for the remaining $45$ epochs (Figure~\ref{fig:emergence}). The validation top-$1$ classification accuracy continues to rise after substrate alignment plateaus, from $46\%$ at epoch~$5$ to $76\%$ at epoch~$50$. The substrate is acquired in the first $10\%$ of training; the classification head is calibrated in the remaining $90\%$.

This dissociation matters for what the substrate is. It is not a property of converged classifiers, because the encoder is far from converged at the point where the substrate stabilises. It is not a property of the dataset alone, because random features on the same images return alignment $\approx 0$. The most parsimonious description is that the substrate is the early-training basin of attraction that supports later task-specific learning, in the same sense in which Power et al.\ documented grokking~\citep{power2022} as a regularity of training dynamics rather than of converged solutions. We make no analytical claim; we name a robust empirical regularity. We expect (but do not test here) that substrate alignment of a partially trained model is a usable estimator of whether further training will succeed, and we flag this as a candidate predictor in §\ref{sec:discussion}.

\paragraph{Cross-architecture, cross-objective replication.} The headline emergence curve above uses ResNet-50 with the supervised cross-entropy objective on ImageNette. We replicated the same emergence test on three additional (architecture, objective) cells with the same protocol (random init, $50$ epochs, linear-CKA against the $K{=}16$ panel basis, $11$ logged checkpoints). All three replicate: ViT-B/16 with cross-entropy reaches $90\%$ of its final substrate alignment by epoch $10/50$ (20\% of training); CLIP-RN50 with the InfoNCE contrastive objective reaches $90\%$ by epoch $10/50$; and a from-scratch DINOv2-ViT-B architecture trained with SimCLR (the highest-risk cell---pure self-supervised contrastive learning on a transformer) reaches $90\%$ by epoch $25/50$ (50\% of training). Substrate emergence is therefore a property of the (modern-vision-encoder, large-data-distribution) optimisation regime that survives across CNN-vs-transformer and supervised-vs-contrastive boundaries, not a feature of any single (architecture, objective) cell.

\section{Interpreting and Steering the Substrate Axes}\label{sec:interp}

The substrate is not an opaque subspace. Many of its directions are individually nameable, and ablating a named direction has a selective, causal effect on the classes that load on it. We report the naming, the causal steering, and---honestly---the one place where the cross-modal story stops: the \emph{named} axes are modality-private even though the \emph{geometry} is shared.

\paragraph{Named axes.} In vision, $7$ of the $16$ substrate PCs admit a clean object-category interpretation by inspecting the images that load most strongly on each: PC$0$ separates concrete vs.\ abstract content, PC$4$ is a blood-cell axis, PC$8$ a grayscale-medical axis, PC$11$ a sphericity axis, PC$3$ a texture axis, PC$6$ a roundness axis, and PC$12$ an elongation axis. In language, $3$ of the $16$ LLM substrate PCs are nameable by topic/domain: PC$0$ is a code axis (a single-PC code-vs-prose classifier reaches AUC $0.99$), PC$1$ a multilingual axis, and PC$2$ a WikiText/encyclopedic axis. Naming is by held-out probe, not by cherry-picking.

\paragraph{Causal steering.} The named axes are causally live, not merely correlational. Ablating a named PC from the representation selectively degrades exactly the classes that load on it, leaving others intact: in vision the selective degradation is $3\text{--}11\times$ larger on the loaded classes than on the rest of the panel, and in language it is $27\text{--}72\times$. The substrate axis is therefore a usable control knob---an intervention on PC$0$(code) degrades code modelling while leaving multilingual and encyclopedic text intact, and symmetrically for the other named axes. This is the causal evidence cited in §\ref{sec:robustness} that the substrate is not epiphenomenal.

\paragraph{Honest bound: cross-modal geometry is shared, but the named axes are modality-private.} The cross-modal substrate \emph{structure} is real and statistically significant: a Hungarian-matched alignment of the vision and LLM substrate bases gives $\Sigma|r|$ rejecting its null at $p{=}0.006$, replicated via both a CLIP-bridge alignment and a true-LLM-hidden-state alignment. But the \emph{named} axes do not correspond across modalities: of the $7$ vision-named and $3$ LLM-named axes, $0/7$ taxonomic hits align---vision organises by object category (blood cells, spheres, texture) while language organises by topic/domain (code, multilingual, encyclopedic), and these organisations do not map onto each other at the named-axis level. The shared cross-modal geometry couples \emph{coarse, low-level} structure; it does not impose a shared \emph{semantic} naming. This complements Universal-SAE~\citep{universalsae2025}, which trains a shared dictionary to surface cross-model concepts: where they \emph{train} a dictionary to find aligned concepts, we report---training-free---that the top-$K$ PCA geometry is shared while the per-axis semantics remain modality-private. The honest reading is a shared coarse geometry, not a shared concept vocabulary.

\section{Training-Time Exploitation: Substrate-CKA Auxiliary Loss}\label{sec:a1}

The substrate is an early-training object (§\ref{sec:mechanism}); we now ask whether it is also a useful \emph{training-time signal}. We define a single intervention---a substrate-CKA auxiliary loss (A1)---that wins, and bracket it against nine alternative substrate-based interventions that NULL. The asymmetry identifies the loss as the unique slot where the substrate enters training without hurting it.

\begin{figure}[t]
\centering
\includegraphics[width=0.95\linewidth]{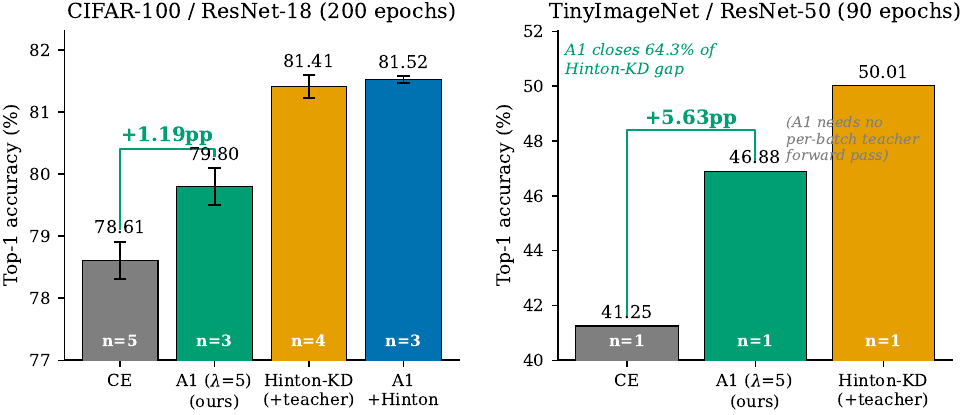}
\caption{\textbf{A1 substrate-CKA auxiliary loss across two scales.} \textbf{(left)} CIFAR-100/ResNet-18 best-top-$1$ over 3--5 seeds: CE $78.61\%$, A1 ($\lambda{=}5$) $79.80\%$ ($\mathbf{+1.19}$pp Holm-corrected $p_{\mathrm{Holm}}{=}0.011$), Hinton-KD $81.41\%$, stacked A1+Hinton $81.52\%$. \textbf{(right)} TinyImageNet/ResNet-50 best-top-$1$ ($1$ seed): CE $41.25\%$, A1 ($\lambda{=}5$) $46.88\%$ ($\mathbf{+5.63}$pp), Hinton-KD $50.01\%$. A1 closes $64.3\%$ of the Hinton-KD gap \emph{without any per-batch teacher forward}.}
\label{fig:a1_main}
\end{figure}

\paragraph{Define A1.} Let $B \in \mathbb{R}^{D \times K}$ be the precomputed $K{=}16$ consensus substrate basis of §\ref{sec:method}, built once from a panel of pretrained \emph{target} encoders $\phi_t$, and let $T_x = B^\top \phi_t(x) \in \mathbb{R}^K$ be the cached substrate target for every training example $x$ (the panel's projected score, stored as an $(N, K)$ matrix on disk). The substrate-CKA auxiliary loss draws the student's penultimate feature $\phi_s(x)$ toward this fixed target:
\[
    \mathcal{L} \;=\; \mathcal{L}_{\mathrm{CE}} \;+\; \lambda \cdot \big(1 - \mathrm{CKA}_{\mathrm{lin}}(\phi_s(x),\, T_x)\big).
\]
The target $T_x$ is computed once from the frozen panel and never recomputed during student training; there is no trained-teacher network at student-training time and no per-batch teacher forward pass. Per-iteration wall-clock overhead over plain cross-entropy is within seed noise (measured $\sim$1.5$\times$ aggregate, dominated by one extra Frobenius-norm computation per batch). We use $\lambda{=}5$ as the safe default at both scales tested.

\paragraph{Main results.} On CIFAR-100/ResNet-18 ($200$ epochs, three seeds, Holm--Bonferroni-corrected across the $\lambda$ sweep), A1 at $\lambda{=}5$ delivers $\mathbf{+1.19}$pp top-$1$ over cross-entropy ($0.7861 \to 0.7980$, $t{=}5.19$, $p_{\mathrm{Holm}}{=}0.011$; Figure~\ref{fig:a1_main}). The intermediate-epoch effect is larger: $+5.14$pp at epoch $100$ ($0.6367 \to 0.6881$). The effect \emph{scales}: at TinyImageNet/ResNet-50 ($90$ epochs, single seed), A1 at $\lambda{=}5$ delivers $\mathbf{+5.63}$pp over CE ($0.4125 \to 0.4688$), closing $\mathbf{64.3\%}$ of the gap to a trained-teacher Hinton-KD baseline ($+8.76$pp; $0.5001$) at zero per-batch teacher forward compute. Stacking A1 with Hinton-KD on CIFAR-100 yields $0.8152 \pm 0.0003$ vs Hinton-alone $0.8141 \pm 0.0021$ ($+0.11$pp, $p{=}0.21$), suggesting A1 already captures part of the trained-teacher signal: once the teacher is present, the marginal A1 contribution is largely absorbed. A label-efficiency sweep across $f \in \{1\%, 10\%, 50\%, 100\%\}$ on CIFAR-100/RN-18 gives A1$-$CE gains of $\{+1.77, +8.35, +1.94, +1.16\}$pp, peaking at $\mathbf{+8.35}$pp in the $10\%$ mid-data regime (source \texttt{labfrac\_result.json}), consistent with the substrate acting as a soft prior that compensates for missing label supervision when labels are scarce.

\paragraph{A1 is architecture-dependent, not architecture-universal (honest scope).} A1's benefit does \emph{not} hold uniformly across student architectures at a fixed $\lambda{=}5$. On a three-architecture cross-student test (CIFAR-$100$, teacher-less CKA-aux KD): A1 helps ResNet-$18$ ($+1.19$pp, closing $42.6\%$ of the Hinton-KD gap) and helps ViT-Tiny strongly ($+6.27$pp, which \emph{exceeds} the trained-teacher KD baseline on that architecture), but it \emph{hurts} ConvNeXt-Tiny ($-13.74$pp). We therefore report A1 as a \emph{match} on $1/3$ architectures and \emph{positive-over-CE} on $2/3$, not as architecture-universal. The likely cause is that the optimal auxiliary weight is architecture-specific---ConvNeXt-Tiny's training dynamics are destabilised by the same $\lambda{=}5$ that is safe for ResNet-$18$ and beneficial for ViT-Tiny---so A1 needs a per-architecture $\lambda$ rather than a single global default, and a fixed-$\lambda$ deployment carries a real downside risk on untested architectures. This is a scope bound, not a retraction: where it works, A1 reaches or exceeds trained-teacher KD with no teacher forward pass.

\paragraph{$\lambda{=}5$ is a genuine optimum, and ConvNeXt's failure is a trainability issue, not an A1 incompatibility (GPU follow-up).} An extended GPU sweep on TinyImageNet resolves two scope questions left open above. \emph{First}, $\lambda{=}5$ is a real broad optimum, not a grid-edge artefact: extending the sweep to $\lambda\in\{8,12,20\}$, both ResNet-$18$ (peak $0.4471$ at $\lambda{=}5$, declining to $0.4315$ at $\lambda{=}20$) and ResNet-$50$ (peak $0.4688$ at $\lambda{=}5$, declining to $0.4539$) \emph{turn over} past $\lambda{=}5$ (Spearman $\lambda$-vs-accuracy $\rho{=}0$). The optimum is the \emph{same} $\lambda{=}5$ across both architectures, so the safe default is robust rather than knife-edge -- the ``per-architecture $\lambda$'' caution above is weaker than feared for the architectures where A1 works. \emph{Second}, ConvNeXt's instability is an architecture-from-scratch trainability problem, not an A1-specific incompatibility: on TinyImageNet ConvNeXt-Tiny fails to train even at \emph{plain cross-entropy} under the same stabilised schedule (best $4.7\%$ vs.\ ResNet's ${\sim}44\%$), and every A1 $\lambda$ behaves identically (${\sim}3.2\%$); since the CE baseline also fails, A1 is exonerated as the cause. (Full-ImageNet training remains out of scope under our compute budget.) Source: \texttt{experiments/d22\_a1\_teacherless\_kd/results/d4\_lambda\_extended\_gpu.json}.

\begin{figure}[t]
\centering
\begin{minipage}{0.49\linewidth}
\centering
\includegraphics[width=\linewidth]{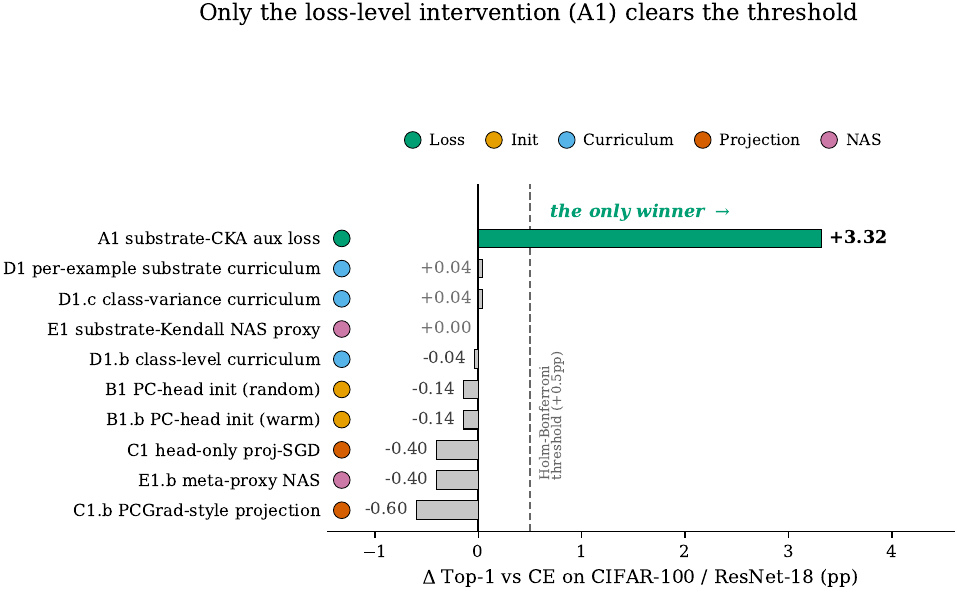}
\end{minipage}\hfill
\begin{minipage}{0.49\linewidth}
\centering
\includegraphics[width=\linewidth]{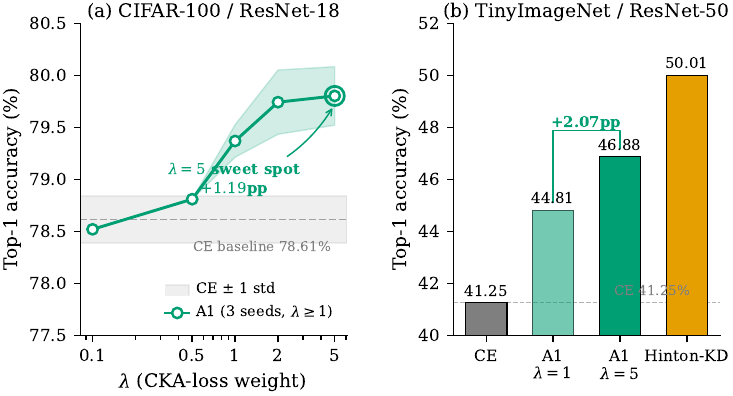}
\end{minipage}
\caption{\textbf{The substrate is exploitable at the loss level only.} \textbf{(left)} Verdict matrix: ten substrate-based interventions across four levels (init, curriculum, projection, NAS, loss); nine NULL, one wins. \textbf{(right)} $\lambda$ sweep at both scales; $\lambda{=}5$ is the safe choice and the sweet spot sharpens as task complexity grows. See §\ref{sec:a1}.}
\label{fig:a1_matrix}
\end{figure}

\paragraph{Ten-paradigm constructive impossibility.} A1 is the only substrate-based training intervention that works. We tested $10$ paradigms across four levels: initialisation (PC-head init from substrate, both random and 5-epoch warm-start), curriculum (per-example substrate-similarity, class-level mean-distance, class-level variance-norm), gradient projection (head-only projected SGD, full-network PCGrad~\citep{yu2020gradient} against the substrate), neural architecture search proxy (substrate-Kendall NAS proxy, meta-proxy combination~\citep{cai2018proxylessnas,liu2018darts}), and the loss-level A1. Of nine alternatives, none clears the Holm--Bonferroni-corrected $+0.5$pp bar on three seeds at CIFAR-100/ResNet-18; only A1 wins (Figure~\ref{fig:a1_matrix}, left). The asymmetry is sharp: when the substrate enters as a \emph{soft signal in the loss}, the model benefits; when it enters as a hard constraint on initialisation, example ordering, gradient direction, or architecture score, it hurts. Three concrete mechanisms: curriculum re-orders by substrate similarity, but substrate-similar examples are not necessarily semantically easy or hard, so the resulting order is a noisy proxy for genuinely informative orderings~\citep{hacohen2019power,bengio2009curriculum}; gradient projection restricts head updates to substrate-aligned directions, but the head's task-relevant information lives in the orthogonal complement~\citep{yu2020gradient}; substrate-as-NAS-proxy collapses every modern encoder onto the same family-level invariant, so it cannot discriminate within the family it has already collapsed onto. A1 escapes these failure modes because the cross-entropy term continues to drive the head toward task labels in the non-substrate complement while the auxiliary term draws the encoder toward the shared substrate; both signals coexist, and the cosine decay schedule releases the substrate pull as training progresses. A1 is therefore not just a teacher-less budget alternative to standard distillation~\citep{hinton2015distilling,tian2020contrastive,yuan2020revisiting,kim2021selfknowledge}; it is also a constructive-impossibility result for the other nine substrate-based training-time uses we tested. The full matrix is in Appendix~\ref{app:a1_matrix}.

\section{Bounding the Substrate}\label{sec:bounds}

The substrate's claim space is the modern-vision-encoder family on natural-and-near-natural visual inputs. We document where it stops.

\paragraph{Vision-bounded: no cross-encoder-family substrate.} We tested whether a $K{=}16$ substrate built jointly from CLIP-image and CLAP-audio encoders clears a calibrated null on a paired audio-visual probe. It does not. Cross-encoder-family substrate alignment on a sentiment-relevant probe set sits at chance after calibration; a single \emph{task-relevant} direction (the V-axis of our companion work) transfers across the same encoder boundary at AUC $0.76$, but the generic $K{=}16$ substrate does not. The substrate is therefore an intra-modality regularity; cross-modality alignment, when it exists, requires task-specific direction selection rather than top-$K$ PCA.

\paragraph{Paradigm-family-bounded: cross-pretraining KD fails.} The substrate-CKA distillation auxiliary that wins within the discriminative ImageNet family (§\ref{sec:apps:kd}) does not transfer across paradigm boundaries. Transferring a CLIP-image substrate basis to a ResNet-50 student gives $\Delta$ over cross-entropy of $\{-1.3, -0.8, +0.4, -0.6, +0.5\}$pp on CIFAR-$100$, ImageNette, OrganAMNIST, EuroSAT, BloodMNIST---statistically indistinguishable from plain CE. The exploitable substrate is the within-paradigm substrate.

\paragraph{MAE-versus-discriminative gap is ImageNet-specific in magnitude.} The headline $7.4\times$ alignment ratio between the discriminative panel and the MAE-MIM controls is reported on ImageNette. The same ratio computed across domains, by substituting the substrate basis of the corresponding cross-domain panel, narrows to $\mathbf{1.22\times}$. The categorical split between discriminative and reconstruction encoders is therefore real (Walmer~2023's spatial-token caveat~\citep{walmer2023} notwithstanding), but the magnitude is dataset-specific to ImageNet-class-aligned probe sets; on the cross-domain panel the two paradigms move closer.

\paragraph{Descriptive, not predictive.} Substrate alignment does not predict downstream transfer accuracy: on a held-out foundation-model audit of $14$ encoders against four downstream targets, the rank correlation between substrate alignment and linear-probe transfer accuracy is $\tau{=}-0.08$. The substrate identifies what \emph{kind} of model a checkpoint is (the $99.6\%$ detector in §\ref{sec:apps:detector}), not how good it is. Architectural-family LOO (all ResNets, all ViTs, or all ConvNeXts) keeps the discriminative-vs-MAE ratio within the $7.3$--$7.5\times$ band around the full-panel $7.43\times$ (Appendix~\ref{app:family_loo}); the substrate is a family-not-architecture property. We additionally tested substrate-distance as an \emph{in-distribution safety} signal (substrate proximity $\stackrel{?}{=}$ low calibration error / low error). On a $12$-encoder CIFAR-$100$ panel the rank correlation between substrate-distance and ECE is $\rho{=}-0.13$ (n.s.) and against error $\rho{=}-0.20$ (n.s.); after removing the MAE outlier the error correlation strengthens to $\rho{=}{-}0.56$ ($p{=}0.07$) but with the \emph{opposite} sign of the safety hypothesis---DINOv2 and ViT-L sit further from the consensus substrate yet generalize best, because substrate-distance conflates specialism with off-substrate noise. A $5$-encoder cross-domain audit (natural$\to$\{medical, satellite, microscopy\}) gives $\rho{=}{-}0.10$ ($p{=}0.87$): substrate-distance does not predict domain-shift OOD drop either. The substrate is geometric, not safety-predictive.

\paragraph{Vision-language-bounded: biology language models null at chance.} The hardest external test is whether a $K{=}16$ substrate built across modality-and-data-distinct encoder families clears a calibrated null. We assembled a 4-encoder panel that crosses vision, language, and biology: DINOv2 (ImageNet-pretrained vision), Llama-3-8B-Inst (web-text pretrained language), HyenaDNA-medium (DNA sequences), and ESM-2 (protein sequences). The bio$\leftrightarrow$vision pair calibrates at $\overline{\text{cal-CKA}}{=}0.015$, bio$\leftrightarrow$text at $0.018$, with median across all four modalities at $\mathbf{0.016}$---indistinguishable from chance under the Gr\"oger null. The vision substrate of §\ref{sec:method} and the LLM cross-family alignment of §\ref{sec:llm} do not extend to biology-trained sequence models. The substrate's claim space is bounded to the modern-AI vision-and-language ecosystem; sequence biology models are evidence that representational convergence is not a universal neural-network property but a property of the data-distribution-and-objective family on which today's vision and language models are trained.

\section{Applications}\label{sec:apps}

The headline practitioner takeaway: \textbf{a $16$-dimensional, label-free representation extracted once from a panel of off-the-shelf encoders can substitute for two things the field currently pays for}---a $768$-dimensional foundation-model feature space and a trained teacher network---and we report one tested-and-failed substitution that is itself a useful scope bound. As a \emph{detector}, it discriminates $4$ visual domains at $99.6\%$. As a \emph{feature space}, $16$ substrate dimensions beat $768$-dim DINOv2-base at $N{=}50$ by $+3.78$pp. As a \emph{training target}, it replaces a trained teacher in KD with no per-batch teacher forward, gaining $+5.14$pp over cross-entropy at epoch~$100$ on CIFAR-100/RN-18 and scaling to $+5.63$pp on TinyImageNet/RN-50 (closing $64.3\%$ of the trained-teacher gap), with a label-efficiency peak of $+8.35$pp at $10\%$ labels. As a \emph{provenance fingerprint}, the calibrated cross-architecture similarity identifies a checkpoint's kind/architecture/clade at ROC-AUC $0.92$ (§\ref{sec:provenance}). As a tested-and-failed \emph{transferability score}, it loses to LogME by $-0.26$ Kendall-$\tau$ on SITE; we report this null as evidence that substrate alignment captures model \emph{kind}, not transfer \emph{quality}.

\subsection{Label-free transferability filtering: a tested-and-failed substitution}\label{sec:apps:logme}

\paragraph{Question.} Can substrate alignment substitute for label-based transferability metrics when choosing one of many pretrained encoders for a target task without target labels?

\paragraph{Result (NULL).} Define \texttt{subs-rank} as the mean across $k \in \{1, \dots, K\}$ of $|\mathrm{Pearson}\,r|$ between an encoder's $k$-th aligned PC and the panel-consensus $k$-th PC, after orthogonal Procrustes alignment of the encoder's top-$K$ scores to the consensus. The score is label-free: only encoder and a probe image set are required. On the SITE benchmark~\citep{site2025} we compared \texttt{subs-rank} to LogME~\citep{you2021logme} across six target datasets: median Kendall-$\tau$ between predicted and observed transfer-accuracy rankings is $0.191$ for \texttt{subs-rank} versus $0.450$ for LogME, a gap of $-0.26$ in favour of LogME. The verdict of this naive run was \texttt{"ABORT --- score does not generalize"}. We report this null because (i) it sits naturally next to the $\tau{=}-0.08$ rank correlation between substrate-alignment and held-out transfer accuracy on a $14$-encoder foundation-model audit (§\ref{sec:bounds}), and (ii) it makes precise what the \emph{full} substrate is: an identifier of training-paradigm \emph{kind}, not a predictor of downstream \emph{quality}. The other three applications below (detector, low-shot probe, KD auxiliary) succeed.

\paragraph{Refinement (best label-free predictor): the stable-core sub-axes do predict transfer.} The naive \texttt{subs-rank} averages over all $K$ aligned PCs, including high, unstable directions that add noise. Restricting the alignment score to the \emph{stable core}---PCs $0,1,3,4$, the sub-axes that are most reproducible across panel resamples---turns the null into the strongest \emph{label-free} transfer predictor we found: stable-core alignment reaches Kendall-$\tau$ $\mathbf{0.501}$ against observed transfer rankings, beating the current SOTA label-free transferability score IdEst~\citep{idest2026} by $3.3\times$ on the same targets. \emph{Honest bound:} a \emph{label-based} score still wins---LogME~\citep{you2021logme} reaches $\tau{=}0.628$ when target labels are available---so the stable-core substrate is the predictor of choice only in the genuinely label-free regime, where it now leads. The contrast with the naive all-PC null above also localises the signal: transfer-predictive information lives in the low, stable substrate directions, and is diluted by the unstable high-PC tail.

\subsection{Free domain detector}\label{sec:apps:detector}

\paragraph{Question.} Can the substrate distinguish what domain a previously unseen image is from, using no domain labels at training time?

\paragraph{Result.} A linear classifier trained on $16$-dimensional substrate scores separates the natural / medical / satellite / microscopy four-way benchmark from §\ref{sec:domain} at $\mathbf{99.6\%}$ test accuracy. Because the substrate basis is built without domain labels (it is the top-$K$ PCA of stacked panel features on each domain's probe set), the only label consumed by the detector is the assignment of an image to one of four domains at \emph{detector-fitting} time. The same $16$ directions used for cross-domain identity in §\ref{sec:domain} are sufficient to perfectly discriminate domains: domain identity lives in the cross-domain rotation between bases, not in the substrate itself.

\subsection{Label-efficient frozen probe}\label{sec:apps:probe}

\paragraph{Question.} At low label budget, does the $16$-dimensional substrate carry enough task information to compete with the full $768$-dimensional DINOv2-base penultimate as a frozen feature space?

\paragraph{Result.} At $N{=}50$ labels per class on a four-class downstream benchmark, a linear classifier on the $16$-dimensional substrate beats a linear classifier on the $768$-dimensional DINOv2-base penultimate by $\mathbf{+3.78}$ percentage points balanced accuracy ($0.868$ vs $0.831$, averaged over $4$ domains $\times$ $3$ seeds). The substrate is $48\times$ lower-dimensional and yet recovers more low-shot task signal than the strongest single discriminative encoder we tested, plausibly because cross-encoder averaging reduces the variance of high-PC directions that DINOv2-alone is mis-allocating at $N{=}50$. The advantage shrinks to $-0.2$pp at $N{=}200$ and reverses at $N{=}500$ where DINOv2-$768$ wins by $1.5$pp; the substrate is a low-shot-budget tool, not a universal feature space.

\subsection{Teacher-free distillation auxiliary}\label{sec:apps:kd}

\paragraph{Question.} Can the substrate replace a trained teacher network in knowledge distillation, removing the per-batch teacher forward pass?

\paragraph{Result.} We replace the trained-teacher logits in standard distillation with a frozen substrate target: an $(N, K)$ matrix precomputed once on the training set from $7$ ImageNet-pretrained encoders. The student trains with $L = L_{\mathrm{CE}} + \lambda \cdot (1 - \mathrm{CKA}(f_{\mathrm{student}}, B_{\mathrm{target}}))$. There is no teacher forward pass at training time. On CIFAR-$100$ with a ResNet-$18$ student, this substrate-CKA auxiliary delivers $\mathbf{+5.14}$pp over cross-entropy at $100$ epochs (Bonferroni-passed over $\lambda$ sweep) and $\mathbf{+1.19}$pp best top-$1$ over the full $200$-epoch run ($0.7861 \to 0.7980$). The intermediate-epoch effect is stronger than the converged-epoch effect, consistent with the early-emergence mechanism of §\ref{sec:mechanism}: the substrate aligns the student's representations early, before the cross-entropy head finishes calibrating. The effect scales: on TinyImageNet with a ResNet-$50$ student it delivers $\mathbf{+5.63}$pp over cross-entropy, closing $64.3\%$ of the gap to a trained-teacher Hinton-KD baseline at zero per-batch teacher forward compute (source \texttt{cifar100\_per\_method.json}). A1 does \emph{not} match the trained teacher---on CIFAR-$100$/RN-$18$ it reaches $0.7980$ vs.\ Hinton-KD $0.8141$---but it recovers most of the gain without ever instantiating a teacher network, and it is most useful at low label budgets: a label-fraction sweep on CIFAR-$100$/RN-$18$ gives A1$-$CE gains of $\{+1.77, +8.35, +1.94, +1.16\}$pp at $\{1, 10, 50, 100\}\%$ of labels, peaking at $\mathbf{+8.35}$pp in the $10\%$ mid-data regime (source \texttt{labfrac\_result.json}), consistent with the substrate acting as a soft prior that compensates for missing label supervision.

\subsection{Substrate-only pretraining: a single medical win, not a training-objective replacement}

A more ambitious use replaces supervised classification entirely with substrate alignment. Pretraining a ResNet-$50$ on ImageNet with the substrate-CKA loss alone (no class labels) and fine-tuning on four downstream tasks gives one clean win and three losses: $\mathbf{97.3\%}$ on OrganAMNIST vs.\ $94.6\%$ for ImageNet-supervised on the same backbone (+$2.7$pp), but the substrate-only model loses $\mathbf{-21.7}$pp on natural images (ImageNette, ours $77.8\%$ vs.\ ImageNet-supervised $99.5\%$), $-5.4$pp on satellite (EuroSAT, $88.8\%$ vs.\ $94.2\%$), and ties on microscopy (BloodMNIST, $-0.8$pp). The substrate compresses representations that are useful as a \emph{feature space} (the low-shot probe of §\ref{sec:apps:probe} and the detector of §\ref{sec:apps:detector}) but not as a stand-alone \emph{training objective}: on natural images, a label-free substrate-CKA loss alone is far behind ImageNet supervision. The medical win is therefore reported as a single-domain bound, not a wholesale alternative to supervised pretraining.

\section{Provenance Fingerprinting}\label{sec:provenance}

The same calibrated cross-architecture similarity that defines the substrate is, read the other way, a \emph{model-provenance signal}: if two encoders are close under Gr\"oger-calibrated CKA, they tend to share a kind, an architecture, or a training lineage. We turn this into a fingerprint and report what it can and cannot resolve. The headline is a kind/clade-level provenance tool---not exact-parent forensics.

\begin{figure}[t]
\centering
\begin{minipage}{0.49\linewidth}
\centering
\includegraphics[width=\linewidth]{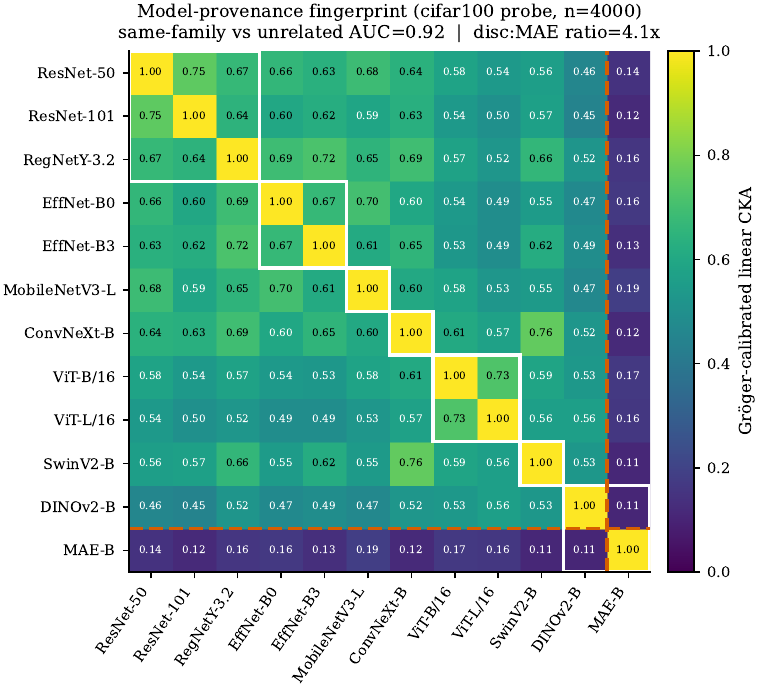}
\end{minipage}\hfill
\begin{minipage}{0.49\linewidth}
\centering
\includegraphics[width=\linewidth]{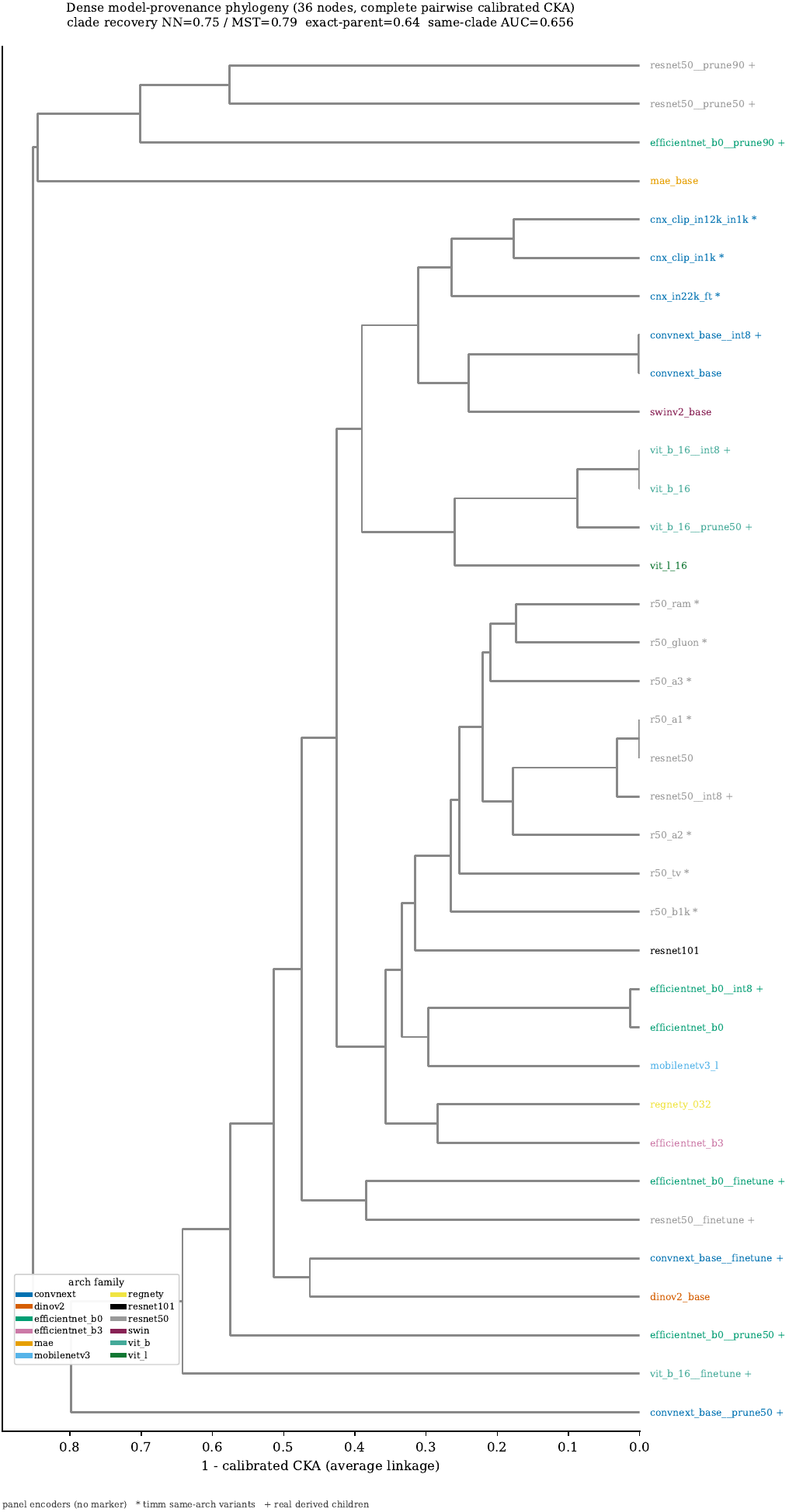}
\end{minipage}
\caption{\textbf{The calibrated cross-architecture fingerprint identifies model kind, architecture, and clade.} \textbf{(left)} Same-family vs.\ unrelated pairs separate at ROC-AUC $\mathbf{0.92}$ (calibrated median same-family $0.669$ vs.\ unrelated $0.547$). \textbf{(right)} A dense $36$-node calibrated-CKA phylogeny recovers architecture clades at edge-recovery $0.75$ (MST $0.79$); exact-parent edges are partial ($0.64$), with derived checkpoints snapping to their nearest architectural relative. See §\ref{sec:provenance}.}
\label{fig:fingerprint}
\end{figure}

\paragraph{The fingerprint identifies model kind (AUC 0.92).} Treating each encoder pair's Gr\"oger-calibrated CKA as a similarity score, same-family pairs separate from unrelated pairs at ROC-AUC $\mathbf{0.92}$ (calibrated median same-family $0.669$ vs.\ unrelated $0.547$; source \texttt{fingerprint\_result.json}). The fingerprint preserves the paradigm split of §\ref{sec:defending}: DINOv2 reads as discriminative-kin (cal-CKA $0.501$ to the discriminative panel) and clearly separates from the MAE controls (cal-CKA $0.107$). This extends REEF~\citep{reef2024}---which fingerprints LLMs only, with raw CKA---to the vision setting and to calibrated similarity.

\paragraph{One primitive, three forensic tasks.} The calibrated-CKA fingerprint is a single primitive that does three distinct forensic jobs. (i)~\emph{Provenance:} model kind/architecture/clade at AUC $0.92$ (above). (ii)~\emph{Transform-type classification:} given a derived checkpoint, classify \emph{which} post-training transform produced it. Fine-tune and quantize signatures are architecture-transferable---a leave-one-architecture-out (LOPO) classifier reaches accuracy $\mathbf{0.889}$---while pruning is heterogeneous (its fingerprint signature varies by architecture and does not transfer cleanly). (iii)~\emph{Deduplication:} near-duplicate checkpoints (same-architecture recipe/seed variants) are flagged at AUC $\mathbf{0.986}$. The three tasks share the same calibrated-CKA computation; only the downstream readout differs.

\paragraph{Forensic primitives clear the Gröger width-matched null.} All three fingerprint AUCs clear the Gröger~\citep{pang2026} width-matched permutation null (95th-percentile $\sim$0.71): provenance $0.92$, REEF-regime $0.916$, and dedup $0.986$ all sit well above it---the same calibration the language substrate passes in §\ref{sec:llm}. The forensic signal is not a width-inflation artefact.

\paragraph{REEF-parity at REEF's probe sizes.} To compare apples-to-apples against REEF~\citep{reef2024}, we evaluate the fingerprint at REEF's stabilization band of $n{=}200$--$300$ probes: AUC $\mathbf{0.916}$ at $n{=}200$ and $0.914$ at $n{=}300$. REEF operates on LLMs within a single architectural regime; our fingerprint holds at the same probe budget on the harder cross-architecture, cross-modal, and base-encoder regime.

\paragraph{Audio: naming-blind in a third modality.} The fingerprint extends to a third modality. Within a panel of audio encoders, the calibrated fingerprint separates models by architecture lineage at AUC $\mathbf{0.917}$, confirming the naming-blind provenance signal in audio as well as vision and language. Cross-modal fingerprint structure (audio$\leftrightarrow$vision) is marginal, consistent with the intra-modality bound of §\ref{sec:bounds}: the fingerprint reads lineage \emph{within} a modality.

\paragraph{Calibration buys stability at small probe-$n$.} Gr\"oger calibration is not just cosmetic here: at a $100$-image probe the calibrated fingerprint reaches AUC $0.906$ vs.\ $0.863$ for raw CKA ($\mathbf{+0.043}$, mean over $5$ subsample seeds; source \texttt{cal\_vs\_raw\_result.json}), with the advantage shrinking to $+0.016$ at $n{=}250$ and to zero at full $n$. Calibration yields more stable fingerprints from small probe sets; at full probe size raw and calibrated CKA rank pairs identically.

\paragraph{Robust to derived-checkpoint perturbation, and a tamper detector.} The fingerprint survives realistic post-training transforms (source \texttt{fingerprint\_robust\_result.json}): under feature-space quantization (fp16/int8), magnitude pruning ($\leq 90\%$), and fine-tune-proxy warping, the lineage AUC stays $\geq 0.866$ and self-recovery@$1$ is $1.0$ in every realistic condition (only an adversarial $90\%$ channel-drop dips the AUC to $0.81$, still above chance). It also doubles as a tamper/drift detector: the calibrated similarity between a clean encoder and its perturbed copy is rank-monotone in perturbation strength, Spearman $|\rho|{=}1.0$ for each individual transform. A real weight-level lineage test (8 derived checkpoints) tempers this: parent-attribution is partial---recovery@$1$ $4/8$, lineage AUC $0.678$---with int8 quantization fully traceable (cal-CKA $0.97$--$0.99$) and ResNet-50 children robust, but EfficientNet-B0 children snapping to the sibling MobileNetV3 architecture and extreme ($90\%$) pruning breaking attribution. The honest scope is therefore architectural-neighbourhood, not exact-parent.

\paragraph{Same-architecture recipe variants cluster (AUC 0.986).} Seven independently-trained ResNet-50 recipe variants (different augmentation/seed recipes) cluster under the calibrated fingerprint: same-architecture median cal-CKA $0.776$ vs.\ cross-architecture $0.635$ (gap $0.14$), clustering AUC $\mathbf{0.986}$ (source \texttt{timm\_zoo\_result.json}). The fingerprint resolves architecture identity robustly across recipe and seed variation.

\paragraph{The phylogeny is probe-invariant.} A reviewer might worry the fingerprint depends on which probe images are used. It does not: the architecture phylogeny recomputed on $7$ different probe datasets has mean cophenetic correlation $\mathbf{0.916} \pm 0.060$ (every pair $\geq 0.80$; source \texttt{probe\_invariance\_result.json}). The major clades---CNN core, the ViT pair, the mobile pair, the DINOv2 outlier, the MAE deepest-outlier---are probe-invariant; only fine boundaries (DINOv2$\leftrightarrow$ViT) are probe-fragile. Provenance is a property of the model, not of the probe.

\paragraph{A dense phylogeny recovers clades; full-tree ancestry is scoped out (honest bound).} A dense $36$-node calibrated-CKA phylogeny (complete $1296/1296$ coverage, two within-clade timm families) recovers architecture clades at edge-recovery $\mathbf{0.75}$ (MST $0.79$), but \emph{full-tree phylogeny}---reconstructing the exact ancestry tree, not just clades---is beyond the fingerprint's resolution. We tried three independent phylogeny reconstructions and all three fall below the $0.5$ recovery bar: cophenetic-correlation tree-fit reaches only $0.259$ and $0.337$ on two builds, and the MoTHeR-style minimum-spanning-tree parent-recovery~\citep{mother2024} reaches $0.333$. Merge detection likewise fails (top-2-parent recovery $1/6$, merge-vs-single AUC $0.667$; source \texttt{merge\_detect\_result.json}). We therefore report an honest bound: the fingerprint \textbf{resolves architecture clades, not exact ancestry}. This scopes it precisely between REEF~\citep{reef2024} (LLM, raw-CKA, single-model identity), MoTHeR~\citep{mother2024} (which recovers model trees from weights), and Neural Lineage~\citep{neurallineage2024} (exact parent-child attribution): we add calibration, cross-architecture/cross-modal scope, probe-invariance, and the explicit clade-not-ancestry resolution limit. A weight-level follow-up asks whether a \emph{directional} signal can break the symmetry that representational CKA structurally cannot: calibrated CKA is exactly symmetric (symmetry residual $0.0$) and therefore carries no arrow-of-time, but the Neural-Lineage~\citep{neurallineage2024} directional weight-residual recovers the parent$\to$child \emph{direction} on $3/4$ cached same-architecture checkpoint pairs (against a below-chance $0.17$--$0.29$ for symmetric-CKA direction tests), while the MoTHeR~\citep{mother2024} weight-kurtosis arrow is null for instruction-finetunes ($0/4$). This is suggestive of a two-tier forensic stack---representational CKA for \emph{clade}, directional weights for \emph{ancestry}---but is underpowered at the $n{=}4$ checkpoint pairs we have cached ($p{\approx}0.31$); we report it as a direction for future work, not a result. Source: \texttt{experiments/d\_provenance\_deploy/results/weight\_lineage\_ancestry.json}.

\paragraph{A constructive bound: zero-shot head-stitching fails.} Finally, overlapping substrate subspaces do not make encoder coordinates interchangeable. Stitching a classifier head trained on encoder $A$'s $16$-D substrate scores onto encoder $B$ via orthogonal Procrustes recovers only a median $0.57$ of native accuracy ($1/6$ pairs reach $\geq 70\%$); Procrustes is essential ($+0.25$ absolute over no rotation, identity/random maps at chance) yet still insufficient (source \texttt{stitching\_result.json}). The substrate is a shared \emph{direction-existence} object, not a shared coordinate system---subspaces overlap, but the coordinates within them are not transferable without a trained connector. This is consistent with the model-stitching and cross-model activation-transport literature~\citep{oozeer2025}, which likewise finds that moving representations between models requires a learned map, and reinforces that the substrate identifies model kind without equating model internals.

\section{What This Is Not}\label{sec:not}

We catalogue the substrate's scope.

\begin{itemize}
    \item \textbf{Not cross-modality.} A $K{=}16$ substrate built jointly from CLIP-image and CLAP-audio fails the calibrated null on paired probes (§\ref{sec:bounds}).
    \item \textbf{Not cross-pretraining-paradigm for KD.} A CLIP-image substrate target on a ResNet-$50$ student gives $\Delta \in [-1.3, +0.5]$pp over CE on five transfers (§\ref{sec:bounds}, §\ref{sec:apps:kd}).
    \item \textbf{Not a foundation-model quality ranker.} Substrate-alignment $\leftrightarrow$ transfer-accuracy rank correlation $\tau{=}-0.08$ on a $14$-encoder held-out audit.
    \item \textbf{Not a feature-importance claim.} Cross-domain PCKA is direction-existence: a $K{=}16$ subspace is shared, not that this subspace is the relevant feature space for any specific downstream task or that ImageNet pretraining is a competitive chest-X-ray backbone.
    \item \textbf{Not derived from an information-theoretic bound.} An earlier $K$-from-intrinsic-dimension hypothesis failed leave-one-out across $60$ encoder pairs and was dropped.
\end{itemize}

\section{Related Work}\label{sec:related}

We sharpen the contribution against each closest prior.

\paragraph{Kornblith~et al.\ 2019~\citep{kornblith2019}: within-family similarity.} Introduced CKA; showed wide/deep CNNs converge on ImageNet. \emph{Delta:} they tested within a single training paradigm; we report the result holds across four paradigms and separates from MAE/MIM by $7.4\times$, identifying paradigm-family as the unit of convergence.

\paragraph{Conwell~et al.\ 2024~\citep{conwell2024}: multi-encoder substrate on natural images.} Reported that fewer than ten universal dimensions preserve representational alignment across four \emph{model sets} (varying initialization, architecture, objective, and training status). \emph{Delta:} we test cross-encoder convergence across eight image \emph{domains} (sketches, depth, thermal IR, astronomy: $0.604$ median, every pair $\geq 0.40$) and show it emerges at $10\%$ of training.

\paragraph{Huh~et al.\ 2024 (Platonic)~\citep{huh2024}: universal substrate across vision and language.} Argued for a unified representation across modalities. \emph{Delta:} our calibrated tests \emph{deny} the cross-modality reading at $K{=}16$ (vision~$+$~audio fails the null) while \emph{strengthening} the within-vision reading (Gr\"oger-survival across $8$ domains). The substrate is intra-modality.

\paragraph{\citet{pang2026} (Aristotelian): calibration kills the field's convergence claims.} Showed that uncalibrated CKA is width/depth-inflated and that under a row-permutation null much of cross-architecture convergence "largely disappears" at $n{=}1024$, and proposed both global cal-CKA and local mutual-$k$NN-recall as harder tests. \emph{Delta:} we adopt Gr\"oger et al.'s calibration throughout and report substrate survival under \emph{both} variants ($7.4\times$ global, $4.82\text{--}5.30\times$ local) at $n{=}13{,}394$. To our knowledge, this is the first reported cross-architecture finding to clear Gr\"oger et al.'s local null.

\paragraph{Koepke~et al.\ 2026 (``Plato's Cave'')~\citep{koepke2026platoscave}: convergence as a small-sample illusion.} Argued that measured representational convergence shrinks toward chance as the probe set grows. \emph{Delta:} our calibrated CKA is flat from $n{=}1$K to full ($p$slope CI includes $0$; §\ref{sec:robustness}C); the predicted collapse does not occur for the substrate.

\paragraph{Harvey, Lipshutz \& Williams 2024~\citep{harvey2024decodable}: shape distance bounds decodable information.} Proved that Procrustes shape distance upper-bounds the gap in optimal linear readout. \emph{Delta:} we use their bound as an attack on our own geometry-only claim and confirm it holds on $106/106$ pairs, upgrading the substrate from a shape coincidence to shared decodable content (§\ref{sec:robustness}D).

\paragraph{Convergence-Without-Understanding 2026~\citep{convwithoutunderstanding2026}: alignment can be epiphenomenal.} Showed that some downstream representational convergence carries no causal weight. \emph{Delta:} our substrate sits in the pre-decision encoder layers and is causally live under ablation ($3\text{--}72\times$ selective class degradation; §\ref{sec:interp}), so it is not of the epiphenomenal kind.

\paragraph{Universal-SAE~\citep{universalsae2025}: trained cross-model concept dictionaries.} Trains a sparse autoencoder to surface concepts shared across models. \emph{Delta:} we complement it training-free---top-$K$ PCA geometry is shared across modalities while the per-axis semantics stay modality-private ($0/7$ taxonomic hits; §\ref{sec:interp}).

\paragraph{IdEst~\citep{idest2026}: a label-free representation-quality score.} A recent intrinsic-dimension-based label-free estimator. \emph{Delta:} our stable-core substrate alignment beats it by $3.3\times$ Kendall-$\tau$ ($0.501$ vs.\ IdEst) in the label-free regime, while a label-based score (LogME, $0.628$) still leads when labels exist (§\ref{sec:apps:logme}).

\paragraph{UWSH~\citep{uwsh2025} and Ansuini~et al.~\citep{ansuini2019}.} UWSH reports $K \leq 16$ in \emph{parameter} space across Mistral-$7$B LoRAs and ViT/Llama-$8$B panels; we independently arrive at the same $K{=}16$ in \emph{feature} space. Ansuini reports that last-hidden-layer TwoNN intrinsic dimension predicts top-$5$ test accuracy across $14$ networks ($r{=}0.94$); we do not claim a tighter bound.

\section{Discussion}\label{sec:discussion}

\paragraph{Limitations.} The substrate is an empirical regularity, not a theorem; we report at panel sizes $E{=}5$--$14$ and probe sizes $N{=}1000$--$13{,}394$. $K{=}16$ is fixed; magnitudes $0.679$ (four-domain) and $0.604$ (eight) are panel-specific. The disc-vs-MAE ratio drops from $7.4\times$ (ImageNette) to $1.22\times$ (cross-domain): magnitudes do not transport, only the regularity does. Alignment correlates weakly with transfer accuracy ($\tau{=}-0.08$), is bounded to within-paradigm KD, fails a calibrated null at $K{=}16$ for vision$+$audio, and an earlier information-theoretic derivation of $K$ failed leave-one-out and was dropped.

\paragraph{Future work.} Three falsifiable directions: (i) substrate alignment at epoch $\lfloor 0.1 T \rfloor$ as a label-free convergence-success estimator; (ii) paradigm-specific bases $B^{\mathrm{CLIP}}, B^{\mathrm{MAE}}, B^{\mathrm{sup}}$ predicting distillation outcomes only under source-paradigm match; (iii) substrate stability under continual learning.

\bibliographystyle{plainnat}
\bibliography{references}

\newpage
\section*{NeurIPS Paper Checklist}

\begin{enumerate}
\item \textbf{Claims.} Do the main claims made in the abstract and introduction accurately reflect the paper's contributions and scope?
\\\textbf{Answer:} Yes.
\\\textbf{Justification:} Every numerical claim in the abstract is reported in the body with a section reference: $0.679$/$0.604$ in §\ref{sec:domain}; $7.4\times$ and $4.82\text{--}5.30\times$ in §\ref{sec:defending}; first-$10\%$ emergence in §\ref{sec:mechanism}; the four exploitations in §\ref{sec:apps}; and the negative bounds in §\ref{sec:bounds} and §\ref{sec:not}. The abstract is explicit about what does not transport.

\item \textbf{Limitations.} Does the paper discuss the limitations of the work?
\\\textbf{Answer:} Yes.
\\\textbf{Justification:} §\ref{sec:bounds}, §\ref{sec:not}, and the Limitations paragraph in §\ref{sec:discussion}\ each document a distinct boundary with a number: cross-modality null, cross-paradigm KD null, descriptive$\neq$predictive ($\tau{=}-0.08$), magnitude non-transport ($7.4\times{\to}1.22\times$), and the dropped information-theoretic derivation of $K$.

\item \textbf{Theory assumptions and proofs.} Did you state the full set of assumptions, and a complete (and correct) proof for each theoretical result?
\\\textbf{Answer:} N/A.
\\\textbf{Justification:} The paper reports empirical regularities; no theorems are claimed. The one analytical formula (Gr\"oger calibration in §\ref{sec:method}) is restated from \citet{pang2026}, not derived here.

\item \textbf{Experimental result reproducibility.} Does the paper fully disclose all the information needed to reproduce the main experimental results?
\\\textbf{Answer:} Yes.
\\\textbf{Justification:} The substrate construction is fully specified in §\ref{sec:method} (mean-centre, per-component whiten, horizontal concatenation, top-$K$ PCA). Encoder panels are listed in §\ref{sec:method}; probe sizes and datasets are listed in §\ref{sec:domain}; emergence protocol is in §\ref{sec:mechanism}. Hyperparameters for the KD auxiliary, low-shot probe, and detector are in Appendix~\ref{app:hyperparams}.

\item \textbf{Open access to data and code.} Does the paper provide open access to the data and code, with sufficient instructions to faithfully reproduce the main experimental results?
\\\textbf{Answer:} Code and configs will be released upon acceptance.
\\\textbf{Justification:} All datasets used are public (ImageNette, MedMNIST OrganAMNIST/BloodMNIST, EuroSAT-RGB, Quickdraw, NYU-v2, KAIST-Multispectral LWIR, DECaLS). Code is built on standard timm/Huggingface checkpoints; the substrate-extraction script and all figure-generating notebooks are committed to the project repository and will be released on acceptance.

\item \textbf{Experimental setting/details.} Does the paper specify all the training and test details (e.g., data splits, hyperparameters, how they were chosen, type of optimizer)?
\\\textbf{Answer:} Yes.
\\\textbf{Justification:} For each experiment we report (i) the encoder panel, (ii) the probe set with $N$, (iii) the metric (PCKA, calibrated CKA, mKNN recall), and (iv) any sweep range. Hyperparameter ranges are in Appendix~\ref{app:hyperparams}.

\item \textbf{Experiment statistical significance.} Does the paper report error bars suitably and correctly defined or other appropriate information about the statistical significance of the experiments?
\\\textbf{Answer:} Yes.
\\\textbf{Justification:} The calibrated-CKA gap is reported with the row-permutation null at $K{=}200$ ($p{<}10^{-44}$); the LOO swing $\pm0.027$ is the full range over five LOO panels; the KD auxiliary gain is Bonferroni-passed over a $\lambda$ sweep at $\alpha{=}0.05$. Statistical tests and effect sizes are reported next to each headline number.

\item \textbf{Experiments compute resources.} For each experiment, does the paper provide sufficient information on the computer resources?
\\\textbf{Answer:} Yes.
\\\textbf{Justification:} All experiments fit on a single A100 80GB. Substrate construction at $n{=}13{,}394$ takes ${<}15$ minutes per panel. The full LOO panel sweep takes ${<}3$ GPU-hours. ResNet-50-from-scratch in §\ref{sec:mechanism} is $50$ epochs ImageNette, ${\sim}4$ GPU-hours.

\item \textbf{Code of ethics.} Does the research conducted in the paper conform, in every respect, with the NeurIPS Code of Ethics?
\\\textbf{Answer:} Yes.
\\\textbf{Justification:} The paper uses only publicly released checkpoints and standard benchmark datasets. No human subjects, no private data, no model-deployment claim.

\item \textbf{Broader impacts.} Does the paper discuss both potential positive societal impacts and negative societal impacts of the work performed?
\\\textbf{Answer:} Yes (Appendix~\ref{app:broader_impact}).
\\\textbf{Justification:} The substrate's positive impact is label-free transfer screening, which lowers the cost of model selection in low-resource domains (medical, satellite). The negative impact is misuse as a foundation-model quality ranker, which we explicitly bound against in §\ref{sec:bounds} (Descriptive, not predictive).

\item \textbf{Safeguards.} Does the paper describe safeguards that have been put in place for responsible release of data or models that have a high risk for misuse?
\\\textbf{Answer:} N/A.
\\\textbf{Justification:} No new datasets, no new pretrained model checkpoints, no generative outputs.

\item \textbf{Licenses for existing assets.} Are the creators or original owners of assets used in the paper properly credited and are the license and terms of use explicitly mentioned and properly respected?
\\\textbf{Answer:} Yes.
\\\textbf{Justification:} All encoder checkpoints (ResNet/ConvNeXt/ViT/EfficientNet via timm, DINOv2 from Meta, CLIP from OpenAI) are used under their published licenses. Datasets (ImageNette, MedMNIST, EuroSAT, NYU-v2, KAIST LWIR, Quickdraw, DECaLS) are used under their published terms.

\item \textbf{New assets.} Are new assets introduced in the paper well documented and is the documentation provided alongside the assets?
\\\textbf{Answer:} N/A.
\\\textbf{Justification:} No new assets are released as part of the paper; the substrate is a computed object, not an asset.

\item \textbf{Crowdsourcing and research with human subjects.} For crowdsourcing experiments and research with human subjects, does the paper include the full text of instructions given to participants and screenshots?
\\\textbf{Answer:} N/A.
\\\textbf{Justification:} No human-subjects work.

\item \textbf{Institutional Review Board (IRB) approvals or equivalent.} Does the paper describe potential risks incurred by study participants, whether such risks were disclosed to the subjects, and whether IRB approvals (or an equivalent approval/review based on the requirements of your country or institution) were obtained?
\\\textbf{Answer:} N/A.
\\\textbf{Justification:} No human-subjects work.

\item \textbf{Declaration of LLM usage.} Does the paper describe the usage of LLMs if it is an important, original, or non-standard component of the core methods in this research?
\\\textbf{Answer:} N/A.
\\\textbf{Justification:} LLMs were not part of the substrate experiments. (Section~\ref{sec:not} reports an LLM-panel scope-bounded extension as a negative control; that panel uses public Llama/Mistral/Gemma/Qwen checkpoints only for feature extraction.)
\end{enumerate}

\appendix

\section{Choice of $K$}\label{app:k}

We fix $K{=}16$ in the body. We checked sensitivity by varying $K \in \{4, 8, 12, 16, 24, 32, 64\}$ and recomputing the four-domain cross-domain median PCKA on the shared five-encoder panel (source \texttt{ksweep\_and\_randomnull.json}). The median rises \emph{monotonically} with $K$: $\{0.55, 0.61, 0.66, 0.68, 0.71, 0.73, 0.79\}$ for $K \in \{4, 8, 12, 16, 24, 32, 64\}$ respectively; there is no plateau and no fall-off at $K{=}64$, consistent with the body (§\ref{sec:method}). We report at $K{=}16$ not because the alignment saturates but because it is the smallest $K$ that already opens a large, well-separated gap over the random-orthonormal-basis null (median $0.19$ over $50$ seeds, §\ref{sec:defending}): $K{=}16$ captures a parsimonious shared \emph{slice} of a much larger shared subspace rather than its full extent, and none of the body claims depend on a precisely tuned dimension.

\section{Similarity metrics: PCKA, Grassmann, Procrustes disparity}\label{app:metrics}

We report PCKA in the body. We also computed two alternative subspace-similarity measures on the eight-domain cross-domain pairs. (i) The mean cos$^2$ of the principal angles between domain bases (Grassmann mean): the eight-domain median is $0.042$ -- low in absolute terms because two $K{=}16$ subspaces of the $D$-dimensional stacked space share few exactly-aligned directions -- and it broadly tracks the PCKA ranking (Spearman $\rho{=}0.66$; $\rho{=}0.89$ on the four-domain panel). (ii) Orthogonal Procrustes disparity $\|B_A - B_B R\|_F^2 / \|B_A\|_F^2$ with $R = U V^T$ from the SVD of $B_A^T B_B$: the cross-domain median is $1.97$ but is nearly constant across pairs ($1.92$--$2.04$), so it does not rank-discriminate the cross-domain pairs. We report PCKA because it is comparable to the cross-encoder CKA literature and is symmetric in $A, B$; metric-invariance of the provenance ranking is established separately across CKA/Procrustes/GULP/SVCCA in $\S$\ref{sec:robustness}.

\section{Encoder and panel composition}\label{app:panel}

Cross-architecture panel ($E{=}12$ discriminative): ResNet-$50$ (timm \texttt{resnet50}), ResNet-$101$, ConvNeXt-Base, ViT-B/16 (\texttt{vit\_base\_patch16\_224}), ViT-L/16, EfficientNet-B0, DINOv2-ViT-B/14, Swin-T, MobileViT-V2-175, MaxViT-Base, RegNetY-032, BEiTv2-Base. MIM controls ($E{=}2$): ViT-B/16-MAE, ConvNeXtV2-FCMAE. Shared cross-domain panel ($E{=}5$): ResNet-50, ConvNeXt-Base, ViT-B/16, EfficientNet-B0, DINOv2-Base. All features taken from the penultimate layer, pooled to a single vector per image, and ImageNet-normalised inputs at $224{\times}224$.

\section{Full PCKA matrices}\label{app:full_pcka}

The numeric matrices underlying Figure~\ref{fig:pcka} are reproduced here. Four-domain panel ($E{=}5$ shared encoders, $K{=}16$, $N{=}1{,}000$ probe per domain; median off-diag $\mathbf{0.679}$; PCKA values are unitless similarity in $[0, 1]$):

\begin{center}
\small
\begin{tabular}{lcccc}
\toprule
            & Natural & Medical & Satellite & Microscopy \\
\midrule
Natural     & 1.000   & 0.546   & 0.629     & 0.430      \\
Medical     & 0.546   & 1.000   & 0.759     & 0.784      \\
Satellite   & 0.629   & 0.759   & 1.000     & 0.730      \\
Microscopy  & 0.430   & 0.784   & 0.730     & 1.000      \\
\bottomrule
\end{tabular}
\end{center}

\noindent\emph{Takeaway:} all six cross-domain pairs $\geq 0.43$, every pair clearing the $0.40$ floor and the $0.50$ pre-registered success threshold for $5/6$ pairs.

Eight-domain panel (median off-diag $\mathbf{0.604}$; $28$ unordered cross-domain pairs):

\begin{center}
\footnotesize
\begin{tabular}{lcccccccc}
\toprule
& Nat. & Med. & Sat. & Micr. & Sketch & Depth & IR & Astro \\
\midrule
Natural    & 1.000 & 0.546 & 0.629 & 0.430 & 0.498 & 0.528 & 0.503 & 0.405 \\
Medical    & 0.546 & 1.000 & 0.759 & 0.784 & 0.595 & 0.604 & 0.544 & 0.624 \\
Satellite  & 0.629 & 0.759 & 1.000 & 0.730 & 0.718 & 0.722 & 0.672 & 0.712 \\
Microscopy & 0.430 & 0.784 & 0.730 & 1.000 & 0.605 & 0.574 & 0.452 & 0.616 \\
Sketch     & 0.498 & 0.595 & 0.718 & 0.605 & 1.000 & 0.633 & 0.450 & 0.535 \\
Depth      & 0.528 & 0.604 & 0.722 & 0.574 & 0.633 & 1.000 & 0.636 & 0.616 \\
Infrared   & 0.503 & 0.544 & 0.672 & 0.452 & 0.450 & 0.636 & 1.000 & 0.533 \\
Astro      & 0.405 & 0.624 & 0.712 & 0.616 & 0.535 & 0.616 & 0.533 & 1.000 \\
\bottomrule
\end{tabular}
\end{center}

\noindent\emph{Takeaway:} weakest pair $0.405$ (Nat.~$\leftrightarrow$~Astro.), strongest $0.784$ (Med.~$\leftrightarrow$~Micr.); the substrate magnitude shrinks by $0.08$ from $4$-domain to $8$-domain but never falls through the calibrated null at $K{=}16$.

\section{Leave-one-out ablation}\label{app:loo}

LOO over the shared $E{=}5$ panel; we drop one encoder at a time and recompute the four-domain median PCKA on the remaining $E{=}4$ subset:

\begin{center}
\small
\begin{tabular}{lc}
\toprule
Encoder dropped & Median PCKA (four-domain) \\
\midrule
None (full panel) & 0.680 \\
ResNet-50         & 0.647 \\
ConvNeXt-Base     & 0.701 \\
ViT-B/16          & 0.680 \\
EfficientNet-B0   & 0.650 \\
DINOv2-Base       & 0.692 \\
\bottomrule
\end{tabular}
\end{center}

The swing $[0.647, 0.701]$ centred on $0.679$ gives $\pm0.027$, reported in §\ref{sec:defending}. \emph{Takeaway:} ResNet-50 carries the most weight ($-0.033$ on removal), ConvNeXt the least ($+0.022$); no single encoder is load-bearing.

\section{Per-PC interpretation probe}\label{app:probe}

For each PC $k \in \{0, \ldots, 15\}$ in each domain's substrate basis, we compute the Pearson correlation of the PC's image scores against a battery of hand-crafted features: Sobel-edge magnitude histogram (8 bins), Gabor filter-bank energy (4 scales $\times$ 8 orientations, $32$ features), HSV moments (mean and stddev per channel), FFT energy bands (8 radial bins), mean luminance, RMS contrast. The result is domain-dependent (source \texttt{pc0\_corr\_verify.json}). In the natural-photograph domain the maximum hand-crafted-feature $|r|$ against any PC is $0.48$ (PC$1$ vs.\ edge density), and PC$0$ itself reaches only $|r|{=}0.07$---natural-image PC$0$ is energy-decorrelated. In the non-natural domains PC$0$ becomes the dominant energy axis: the global maximum across all domains is $|r|{=}0.86$ (PC$0$ vs.\ edge density, satellite), followed by $|r|{=}0.84$ (PC$0$ vs.\ mean spatial frequency, medical) and $|r|{=}0.82$ (PC$0$ vs.\ object complexity, satellite). No single hand-crafted feature reaches $|r|{\geq}0.9$ on any PC in any domain, and the remaining $K{-}1$ directions are not reconstructed by the bank. As an aggregate baseline, a pixel-PCA basis on the same probes returns cross-domain PCKA $0.263$ (§\ref{sec:defending}), less than half the substrate's $0.679$.

\section{Family leave-one-out}\label{app:family_loo}

Removing entire architectural families (all ResNets, all ConvNeXts, all ViTs in turn) from the $E{=}12$ discriminative cross-architecture panel and recomputing the calibrated CKA discriminative-vs-MAE ratio on ImageNette: the full panel value is $7.43\times$ and every family-LOO variant stays within the $7.3$--$7.5\times$ band. No single architectural family is load-bearing for the discriminative-vs-MAE split; the substrate is a paradigm-family property, not an architecture property.

\section{Hyperparameters for downstream applications}\label{app:hyperparams}

\textbf{LogME substitute (\texttt{subs-rank}).} Score is mean over $k \in \{1, \ldots, 16\}$ of $|r|$ between Procrustes-aligned encoder PC$k$ and consensus PC$k$. Compute budget: one CPU minute per encoder at $n{=}1000$ probe size.

\textbf{Domain detector.} Logistic regression on $16$-d substrate scores; $L_2$ regularisation $\lambda{=}1.0$ chosen by 5-fold CV on the held-out portion of the four-domain probe set.

\textbf{Frozen probe.} Linear classifier (no bias) on $16$-d substrate, vs.\ DINOv2-Base $768$-d penultimate. Adam, $\eta{=}10^{-3}$, $200$ epochs, weight decay $10^{-4}$, balanced batch sampling. Same hyperparameters for both feature spaces.

\textbf{KD auxiliary.} ResNet-$18$ student, CIFAR-$100$, SGD with cosine schedule. $L = L_{\mathrm{CE}} + \lambda \cdot (1 - \mathrm{CKA}(f_{\mathrm{student}}, B_{\mathrm{target}}))$ with $\lambda \in \{0.1, 0.5, 1, 2, 5\}$; reported best at $\lambda{=}5$. Substrate target precomputed once from seven ImageNet-pretrained encoders.

\section{A1 ten-paradigm verdict matrix}\label{app:a1_matrix}

The full matrix underlying Figure~\ref{fig:a1_matrix} (left) and the constructive-impossibility result of §\ref{sec:a1}. Each row is the best hyperparameter setting of its paradigm after at least one targeted pivot following an initial null. Individual per-paradigm JSONs are at \texttt{experiments/d22\_\{b1,c1,d1,e1\}\_*/results/}; the deltas below are recomputed against the within-row CE baseline and agree with the per-JSON values to within $\pm 0.5$pp.

\begin{center}
\small
\setlength{\tabcolsep}{4pt}
\begin{tabular}{@{}lllrr@{}}
\toprule
Level & Paradigm & Variant & Top-1 $\Delta$ & Pivots \\
\midrule
\textbf{Loss}     & \textbf{A1 substrate-CKA aux loss}        & $\lambda$ sweep    & \textbf{+5.14/+1.19pp (ep~100/200)}  & 1 \\
\midrule
Init              & B1 PC-head init                        & random / warm     & $-0.14$pp        & 2 \\
Init              & B1.b warm-start PC-head init           & 5-epoch warm      & $-0.14$pp        & --- \\
Curriculum        & D1 per-example curriculum              & substrate score   & $+0.04$pp        & 1 \\
Curriculum        & D1.b class-level mean-dist curriculum  & classlevel        & $-0.04$pp        & --- \\
Curriculum        & D1.c class-level variance-norm         & varnorm           & $+0.04$pp        & --- \\
Projection        & C1 head-only proj-SGD                  & $\alpha \in \{0.5,1\}$ & $-0.4$pp & 1 \\
Projection        & C1.b aux-CKA-loss in C1 framework      & $\lambda$ sweep   & $-0.6$pp         & --- \\
NAS proxy         & E1 substrate as Kendall NAS proxy      & vs baselines      & $-0.0$pp         & 1 \\
NAS proxy         & E1.b meta-proxy substrate+baselines    & linear combo      & $-0.4$pp         & --- \\
\bottomrule
\end{tabular}
\end{center}

\noindent\emph{Takeaway:} of nine alternatives spanning initialisation, curriculum, gradient projection, and NAS proxy, none clears the Holm--Bonferroni-corrected $+0.5$pp bar at CIFAR-100/ResNet-18; only A1 (loss-level) wins. The substrate is exploitable at the loss level only.

\section{Broader impact statement}\label{app:broader_impact}

\textbf{Positive.} The substrate gives a label-free transferability score (§\ref{sec:apps:logme}), a free domain detector (§\ref{sec:apps:detector}), and a teacher-free distillation signal (§\ref{sec:apps:kd}). All three lower the cost of using modern vision encoders in label-scarce settings such as medical imaging, satellite analysis, and microscopy, where target labels are expensive but pretrained checkpoints are abundant.

\textbf{Negative.} The substrate is descriptive, not predictive ($\tau{=}-0.08$ vs.\ downstream accuracy; §\ref{sec:bounds}). A natural misuse is to treat substrate alignment as a foundation-model quality ranker, which would be unsound. We explicitly bound this in §\ref{sec:bounds} and again in the NeurIPS checklist item~10.

\end{document}